\documentclass[lettersize,journal]{IEEEtran}
\usepackage{arydshln} 
\usepackage{amsmath,amsfonts}
\usepackage{array}
\usepackage{textcomp}   
\usepackage{graphicx} 
\usepackage{booktabs}
\usepackage{multirow}
\usepackage{diagbox} 
\usepackage{hyperref}
\usepackage[lined,boxed,commentsnumbered,ruled]{algorithm2e}
\usepackage{bbm, dsfont}
\usepackage{cleveref}
\usepackage[cmyk]{xcolor}
\usepackage{colortbl}
\usepackage{color}
\usepackage[caption=false,font=normalsize,labelfont=sf]{subfig}
\usepackage{graphicx}
\usepackage{float}

\usepackage{subcaption}
\newcommand{\ie}{\textit{i}.\textit{e}., }
\newcommand{\eg}{\textit{e}.\textit{g}., }

\begin{document}
 
\title{Deep Reversible Consistency Learning for Cross-modal Retrieval}

\author{Ruitao Pu*, Yang Qin*, Dezhong Peng\#, Xiaomin Song, Huiming Zheng

\thanks{
R. Pu and Y. Qin are with the College of Computer Science, Sichuan University, Chengdu, China, 610044. 

D. Peng is with the College of Computer Science, Sichuan University, Chengdu, China, 610044, and with Sichuan National Innovation New Vision UHD Video Technology Co., Ltd, Chengdu, China, 610095, and also with National Innovation Center for UHD Video Technology, Chengdu, China, 610095.

X. Song and H. Zheng are with Sichuan Newstrong UHD Video Technology Co., Ltd, Chengdu, China, 610095.

* R. Pu and Y. Qin contributed equally.

\# Corresponding author: Dezhong Peng.
}
 }

\maketitle

\begin{abstract}
Cross-modal retrieval (CMR) typically involves learning common representations to directly measure similarities between multimodal samples. Most existing CMR methods commonly assume multimodal samples in pairs and employ joint training to learn common representations, limiting the flexibility of CMR. Although some methods adopt independent training strategies for each modality to improve flexibility in CMR, they utilize the randomly initialized orthogonal matrices to guide representation learning, which is suboptimal since they assume inter-class samples are independent of each other, limiting the potential of semantic alignments between sample representations and ground-truth labels. 
To address these issues, we propose a novel method termed Deep Reversible Consistency Learning (DRCL) for cross-modal retrieval. DRCL includes two core modules, \ie Selective Prior Learning (SPL) and Reversible Semantic Consistency learning (RSC). More specifically, SPL first learns a transformation weight matrix on each modality and selects the best one based on the quality score as the Prior, which greatly avoids blind selection of priors learned from low-quality modalities. Then, RSC employs a Modality-invariant Representation Recasting mechanism (MRR) to recast the potential modality-invariant representations from sample semantic labels by the generalized inverse matrix of the prior. Since labels are devoid of modal-specific information, we utilize the recast features to guide the representation learning, thus maintaining semantic consistency to the fullest extent possible. In addition, a feature augmentation mechanism (FA) is introduced in RSC to encourage the model to learn over a wider data distribution for diversity. Finally, extensive experiments conducted on five widely used datasets and comparisons with 15 state-of-the-art baselines demonstrate the effectiveness and superiority of our DRCL.  Code is
available at \url{https://github.com/perquisite/DRCL}.
 
\end{abstract}


\section{Introduction}
\IEEEPARstart{T}{he} rapid growth of the Internet has led to an exponential increase in multimedia data~\cite{NUS-WIDE,sun2023hierarchical_uni, sun2024robust}. The data from various modalities or media, such as images, texts, and videos, frequently pertain to similar topics and content, so it is a common practice for analysis to employ the samples from one modality as the query to retrieve related samples in other modalities. However, there is a heterogeneity gap between multimodal data due to inconsistent representation and distribution, which prevents directly measuring the similarity between multimodal data for retrieval. Hence, a method capable of eliminating the heterogeneity gap for cross-modal retrieval is needed.


To overcome this, subspace learning is a prevalent strategy in various methods \cite{KCCA,CCA, sun2024dual,sun2023hierarchical_bi}, in which modal-specific functions are utilized to map multimodal data in a shared subspace, thus eliminating the heterogeneity gap. Among them, traditional methods such as Canonical Correlation Analysis (CCA) \cite{CCA}, aim to maximize statistical similarity between data to learn the shared subspace. However, due to the intricate correlation patterns inherent in real-world data, linear modeling has limited effectiveness in dealing with such data. Although some CCA-variants \cite{KCCA} leverage kernel tricks to capture non-linear relationships between data, determining the most suitable kernel function remains an open challenge.

Due to the powerful nonlinear modeling capabilities of deep neural networks (DNNs), various DNNs-based methods \cite{NACON, HAT, HGAN, MRL, MAN, SDML, MARS, ComqueryFormer} have been proposed to learn the common semantic space.
For instance, Hu \textit{et al.} \cite{DSCMR} introduced a  Deep Supervised Cross-modal Retrieval method (DSCMR) to align intra-class sample representations and make inter-class sample representations more discriminative by utilizing the semantic information in labels. 
Qian \textit{et al.} \cite{GNN4CMR} employed Graph Neural Networks (GNNs) to extract relevant information between labels and leverage dual-adversarial generative networks to eliminate modal-specific information while aligning semantics.
However, these methods are specifically tailored for the data in processing two modalities, which would significantly increase the learning complexity when performing multiple (more than two) modalities interactions, as they require pairwise training to align across multiple modalities.
To address this, some methods specifically target retrieval between multiple modalities by pairwise training simultaneously, such as MRL \cite{MRL} and MAN \cite{MAN}. However, these methods require the multimodal samples to be paired, which might be unavailable or hard due to inconsistent data collection in reality. 
For this issue, some approaches \cite{SDML, MARS} focus on improving flexibility in cross-modal retrieval. For example, Hu \textit{et al.} \cite{SDML} pre-define a common subspace by randomly initializing a mutually orthogonal matrix to guide subspace learning for each modality. Wang \textit{et al.} \cite{MARS} also utilize a randomly initialized and mutually orthogonal matrix to align features from the common subspace and semantics in labels. Simultaneously, a network trained in the first modality is used to generate a feature for each category in the common subspace, guiding subspace learning for each modality. Both methods ensure independent and non-interfering subspace learning for each modality, enhancing flexibility in cross-modal retrieval. 

It is noticeable that the aforementioned methods \cite{SDML, MARS} deploy shared matrices or fully connected layers to align the semantics between sample representations and labels. In addition, they generate a feature for each category as an anchor or prior to guide representation learning for all modalities. To enhance the discriminability between inter-class samples in the common subspace, these methods assume inter-class samples are independent of each other and utilize mutually orthogonal matrices to guide cross-modal learning.  However, according to our observation, using mutually orthogonal matrices to constrain subspace learning is suboptimal as detailed in~\Cref{section: The Impact Of Our Prior On Other Methods} because it fails to take dependencies between inter-class samples into account, limiting the semantic alignment potentiality between sample representations and ground-truth labels. Additionally, MARS \cite{MARS} deploys fully connected layers to generate a feature for each category in the subspace while deploying a matrix to map sample representations into label space. However, there is no guarantee that the two processes are reversible, thus failing to ensure the consistency of the semantics in the conversion.

\begin{figure*}[h]
    \centering
    \includegraphics[width = 0.95\textwidth]{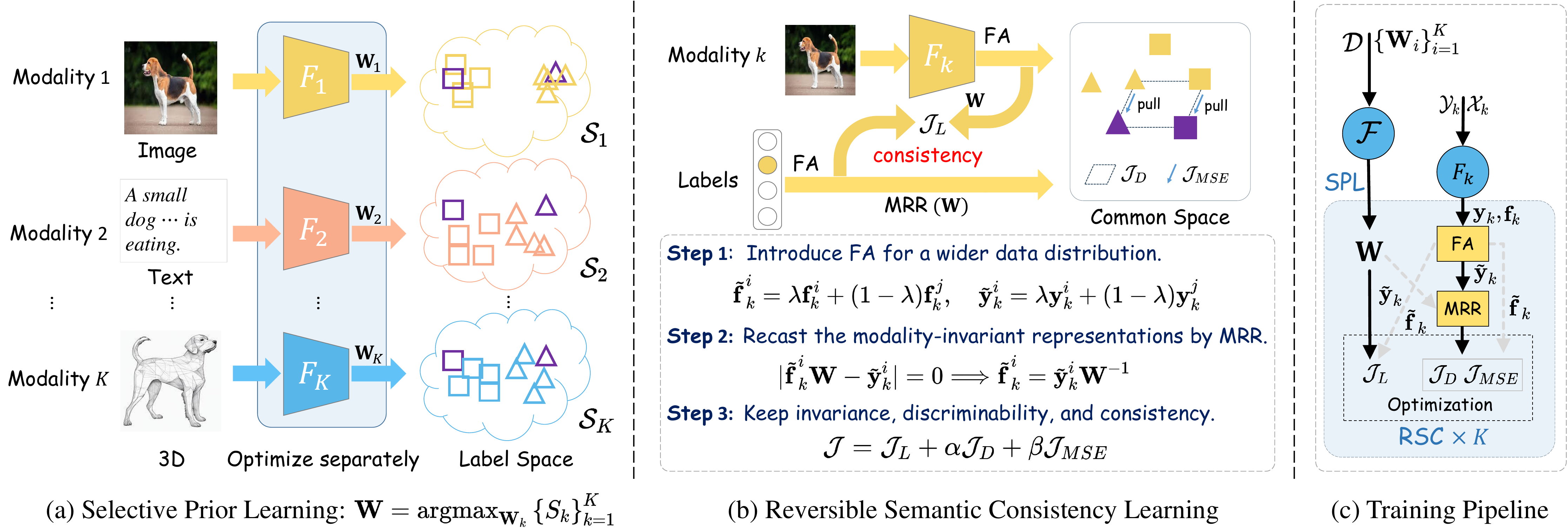}
    \caption{The overall framework of DRCL includes two modules: Selective Prior Learning (SPL) and Reversible Semantic Consistency learning (RSC). (a) In SPL, we first optimize the transformation weight matrices on each modality and select the best one based on the quality scores $\{\mathcal{S}_k\}^K_{k=1}$ as the prior to guide the subsequent RSC. (b) In RSC, we introduce a feature augmentation mechanism (FA) to encourage the model to learn over a wider data distribution. Then, we utilize the Modality-invariant Representation Recasting mechanism (MRR) to recast the modality-invariant representation ($\tilde{\mathbf{y}}^i_k\mathbf{W}^{-1}$) for each semantic category by the prior matrix $\mathbf{W}$ for the subsequent learning. Lastly, we employ $\mathcal{J}_L$ for semantic consistency in the label space, $\mathcal{J}_{D}$ to enhance intra-class compactness and inter-class discriminability, and $\mathcal{J}_{MSE}$ for semantic invariance in the common subspace. (c) is the overall training pipeline of DRCL, including one-time SPL and $K$ times RSC learning modality-specific encoders ($\mathcal{F}=\{F_k\}^K_{k=1}$) for all modalities.}
    \label{fig: framework}
\vspace{-0.3cm}
\end{figure*}

To address the shortcomings mentioned above, this paper introduces a method termed Deep Reversible Consistency Learning (DRCL) for cross-modal retrieval. As illustrated in~\Cref{fig: framework}, the method comprises a Selective Prior Learning (SPL) module and a Reversible Semantic Consistency learning (RSC) module. Specifically, we no longer require shared transformation weight matrices to be mutually orthogonal to guide cross-modal learning. To obtain a more appropriate prior, SPL first optimizes the independent transformation weight matrix on each modality and then selects the best one as the prior for representation learning based on quality scores, which greatly avoids the blind selection of the priors learned from low-quality modality.  Secondly, to ensure that the semantics are always reversible during the conversion process, RSC deploys the generalized inverse matrix of the selected prior matrix to convert sample labels into the corresponding category features as modality-invariant representations. To promote diversity, RSC first employs a feature augmentation mechanism (FA) to encourage the model to learn over a wider data distribution by executing the mixup operation in the embedding space, which is a more flexible way since it does not need to pay attention to the modality format in the input space. To eliminate the cross-modal discrepancy, RSC minimizes the distance between modality-invariant representations and representations of corresponding intra-class samples. Simultaneously, RSC strives to minimize the distance between sample feature distributions and category feature distributions with the same semantics, while maximizing the gap between those with different semantics. This ensures the compactness between intra-class sample representations and the discriminability for inter-class sample representations. Finally, RSC utilizes a label loss to maintain semantic consistency between the sample representations and the actual labels. 
 Overall, the contributions of this paper are outlined as follows:
\begin{itemize}
    \item[(1)] We propose a novel method, namely Deep Reversible Consistency Learning (DRCL), for cross-modal retrieval. DRCL flexibly performs cross-modal learning and eliminates the necessity for pairwise training when involving multiple modalities, thereby greatly reducing training complexity.

    \item[(2)] A Selective Prior Learning (SPL) module is proposed to obtain an ideal prior to guide representation learning, which retains the best characteristics of multi-modal data. We demonstrate that it can be directly used in existing methods ( \eg SDML and MARS) to improve performance.

    \item[(3)] A Reversible Semantic Consistency learning (RSC) module is designed to recast the modality-invariant representation by the Modality-invariant Representation recasting mechanism (MRR) to enhance inter-class discriminability and intra-class compactness while keeping the semantic consistency. 
 
    \item[(4)] We conduct extensive experiments on five benchmark datasets. The results demonstrate that our proposed method outperforms 15 state-of-the-art methods, showcasing its effectiveness. Besides, detailed analysis illustrates the rationality of the design of our method.
        
\end{itemize}

\section{Related Work}
\subsection{Subspace Learning}
Subspace learning~\cite{CCA, jSPSH, 10684088, 9762016, ASFOH}, as a prevalent learning paradigm for cross-modal retrieval, can learn a promising common subspace that accurately measures the similarities between cross-modal samples.  Traditional subspace learning methods project data from different modalities into a shared space by learning modality-specific linear projection matrices. Canonical Correlation Analysis (CCA) \cite{CCA}, Multi-set CCA (MCCA) \cite{MCCA_set}, and Cross-Modality Correlation Propagation (CMCP) \cite{CMCP} exemplify this approach by maximizing the correlation between data to learn linear projection matrices. However, these methods don't leverage semantic information from the labels. In contrast, Generalized Multiview Analysis~\cite{GMA} introduces the Fisher criterion to enhance the discriminative nature of features with different semantics categories in the common subspace. 
ASFOH~\cite{ASFOH} uses asymmetric supervision signal fusion-oriented to improve the discriminability of shared semantic representation. Despite their commendable performance, these methods fall short in capturing highly nonlinear relationships present in real-world data due to their reliance on linear models. Although nonlinear models incorporating kernel tricks, such as kernel CCA~\cite{KCCA}, have been proposed to address this limitation, identifying the appropriate kernel function for specific data remains an open challenge. In response to this, deep learning methods with strong nonlinear modeling capabilities have emerged to capture intricate relationships between data modalities. For example, Cross-Modal Subspace Clustering DCCA~\cite{CMSC-DCCA}, Multi-View Multi-Label CCA~\cite{MVMLCCA}, and Variational Autoencoder with CCA~\cite{VAE-CCA} utilize Deep Neural Networks (DNNs) to project data into a subspace and subsequently capture correlations between modalities.  

However, these methods usually perform pairwise learning for two views/modalities. When multiple modalities are involved, all combinations need to be learned, which brings huge training costs and cannot be well applied to real-world scenarios. In this paper, we focus on utilizing the DNN-based models in an efficient way to achieve cross-modal learning in a common subspace, thus bridging the heterogeneity gap across modalities.

\subsection{Deep Cross-modal Retrieval}
Thanks to the powerful nonlinear modeling capabilities inherent in DNN models, a plethora of approaches  \cite{ NACON, HAT, HGAN, hu2023cross, ODmAP, FNE, CKDH, MLLM, radford2021learning, CL2CM, AlRet, ComqueryFormer} have emerged to address the heterogeneity challenges in cross-modal retrieval (CMR) by various Paradigms, \eg cross-modal hashing~\cite{ hu2022unsupervised, CKDH}, vision language pre-training~\cite{radford2021learning}, multimodal large language model~\cite{MLLM}, and \textit{etc.} In this paper, we focus on category-based cross-modal retrieval. Based on the type of training paradigm, existing deep CMR methods for category-based cross-modal retrieval can be roughly divided into two groups: paired-oriented methods and paired-free methods. The paired-oriented methods \cite{HAT, FNE, AlRet} commonly focus on learning projection networks of different modalities by extracting the semantic co-occurrence information from the same instance pair. On the one hand, to improve discriminability, some methods~\cite{hu2023deep, GNN4CMR, COXI} leverage semantic information in the labels to improve semantic alignments on features in the common subspace. For example, GNN4CMR~\cite{GNN4CMR} employs graph neural networks to capture correlative semantic information between labels, while deploying dual adversarial generative networks to eliminate modality-specific information, facilitating improved semantic alignment. To improve the efficiency, some methods~\cite{MRL, MAN, ELRCMR} jointly train data from multiple modalities. However, these methods require data from all modalities to be paired, which is often unavailable in reality. Unlike paired-free CMR methods \cite{SDML, MARS} featuring flexible learning frameworks that train each modality independently have been proposed and obtained promising results from the well-designed training pipeline. For instance, MARS \cite{MARS} employs a mutually orthogonal matrix for mapping sample representations into label space, simultaneously using a label network learned on the first modality to extract label features devoid of modality-specific information, guiding modality-specific sub-network learning. The above methods independently train data from each modality without mutual interference, effectively reducing computational complexity while maintaining high performance.

Although the previously mentioned methods have made considerable progress, the majority of them can handle only two modalities simultaneously or require paired modal data, which limits real-world application. The paired-free methods adopt flexible learning frameworks to mitigate this challenge, but they overlook dependencies between inter-class samples. In contrast, this paper introduces a method that integrates both considerations of dependencies between inter-class samples and flexibility in processing multiple modalities.

\section{The Proposed Method}
\subsection{Problem Formulation}\label{section:3.1}
 Without loss of generality, we first define some notations to clearly illustrate this work.  Let $\mathcal{D}=\{ \mathcal{X}_k,\mathcal{Y}_k \}^K_{k=1}$ be the dataset with $K$ modalities, where $ \mathcal{X}_k$ means the sample set from the $k$-th modality and $\mathcal{Y}_k$ the corresponding category label set. The samples from the $k$-th modality are represented as $\mathcal{X}_k =\left \{\mathbf{x}_{k}^{1},\mathbf{x}_{k}^{2},\cdots,\mathbf{x}_{k}^{N_k}  \right \}$, where $\mathbf{x}_{k}^{i}$ represents the $i$-th sample from the $k$-th modality and the $N_k$ is the number of samples from the $k$-th modality. The labels from the $k$-th modality are donated as $\mathcal{Y}_k = \left \{ \mathbf{y}_{k}^{1}, \mathbf{y}_{k}^{2},\cdots,\mathbf{y}_{k}^{N_k} \right \}$, where  $\mathbf{y}_k^i\in\mathbb{R}^{1 \times C}$ denotes the label of the sample $\mathbf{x}_{k}^{i}$ and $C$ is the number of categories.  
Note that if $\mathbf{x}_{k}^{i}$ belongs to the $c$-th category $c\in\{1,2,\cdots,C\}$, the $c$-th element of $\mathbf{y}_k^i$ is 1, \ie ${y}_k^{i,c}=1$, otherwise ${y}_k^{i,c}=0$.  
The $k$-th modality-specific encoder that projects samples into the common subspace is defined as $F_k(\cdot,\Phi_k)\in \mathbb{R}^{1\times d}$, where $d$ is the dimensionality of the common subspace. For brevity, $F_k(\mathbf{x}_k^i,\Phi_k)$ is denoted as $\mathbf{f}_k^i$ in the following. The transformation weight matrix of the $k$-th modality is denoted as $\mathbf{W}_k \in \mathbb{R}^{d\times C}$. 

To eliminate the heterogeneity gap between different modalities, we expect the multimodal samples with the same semantics compact and the ones with different semantics far apart in the common subspace. To achieve this, we propose a Deep Reversible  Consistency  Learning (DRCL) framework to constrain cross-modal learning, thus maintaining semantic consistency to the fullest extent possible. Our DRCL consists of two steps: 1) To achieve sample-to-label transformation for preserving semantic constraints, we first learn a transformation weight matrix on each modality and select the best one based on the quality scores as the prior. This process is called Selective Prior Learning (SPL). 2) Then, we employ the reversible matrix of the learned prior to guide the representation learning while keeping the consistency between label space and common space for better invariance, discriminability, and consistency. The process is called Reversible Semantic Consistency learning (RSC). Next, we will elaborate on SPL and RSC for cross-modal retrieval.

 \subsection{Selective Prior Learning}
To alleviate the modality discrepancy,  previous works~\cite{MARS, DSCMR} usually utilize a shared classifier or orthogonal transformation weight matrix to constrain and learn modality invariant features. However, the shared classifier strategy usually requires multiple modalities to be trained jointly, which limits the flexibility of the cross-modal model, especially in processing multiple modalities. In addition, as discussed in~\Cref{sec_pa}, we observe that the orthogonal strategy used in the transformation weight matrix is suboptimal in practical situations since the forced assumption of irrelevances of inter-class samples. To this end, we employ the matrix with the best quality score selected from the transformation weight matrixes trained on all modalities as the prior to guide representation learning, which can globally maintain the semantic consistency between representations and labels to the fullest extent.  To be specific, we optimize an identical transformation weight matrix on each modality and formulate a loss to guide the learning of the transformation weight matrix. For the $k$-th modality,  the loss is defined as follows:
\begin{equation}
         \mathcal{J}_p =\frac{1}{qN_k} \sum_{i = 1}^{N_k}\left (  1-\big(\sum_{c=1}^{C} y_{k}^{i,c} \frac{e^{\mathbf{f}_{k}^{i}\mathbf{w}_k^c}}{ {\textstyle \sum_{j=1}^{C}} e^{\mathbf{f }_{k}^{i}\mathbf{w}_k^j}} \big) ^q\right )  \label{gce_p}
\end{equation}
where $\mathbf{f}_{k}^{i}$ denotes the representation of the sample $\mathbf{x}_k^i$ in the common subspace, $\mathbf{w}_k^c\in\mathbb{R}^{d\times 1}$ is the $c$-th column vector of $\mathbf{W}_k$, and $q\in (0,1]$ is a regulatory factor that controls the hardness. In our experiments, we let $q$ gradually increase from a very small value to 1 with the number of training epochs. The intuition behind this lies in balancing the hardness of learning. In the early stages of training, the smaller $q$ penalizes the semantic alignment deviations more, which helps the model pay more attention to the samples whose semantics are harder to align, allowing the model to learn basic patterns and features faster. As the training progresses, gradually increasing $q$ can lessen the penalty for semantic alignment bias, allowing the model to focus more on overall performance later in training.
 
After obtaining all learned modality transformation weight matrices by $\mathcal{J}_p$, we compare them by a quality score that measures the quality of these weight matrices and select the best one to guide subsequent cross-modal learning.  The score of the matrix $\mathbf{W}_k$ of the $k$-th modality is computed as follows:
\begin{equation}
  \mathcal{S}_k =\frac{1}{N_{k}} \sum_{i = 1}^{N_{k}}\sum_{c=1}^{C} y_{k}^{i,c} \frac{e^{\mathbf{f}_{k}^{i}\mathbf{w}_k^c}}{ {\textstyle \sum_{j=1}^{C}} e^{\mathbf{f }_{k}^{i}\mathbf{w}_k^j}}. \label{score}
\end{equation}
Obviously, the higher the score is, the closer the representations of the samples are to the semantics of the ground-truth labels. Then, we select the learned matrix with the highest score as the prior, \ie $\mathbf{W}=\operatorname{argmax}\limits_{\mathbf{W}_k} \{\mathcal{S}_k\}^K_{k=1} $.

\subsection{Reversible Semantic Consistency Learning}
In this section, we introduce a new cross-modal learning paradigm that learns the modality-invariant and inter-class discriminative representations for cross-modal retrieval, namely Reversible Semantic Consistency learning (RSC).
First, to encourage the model to learn over a wider data distribution, RSC introduces a Feature Augmentation mechanism (FA). Specifically, we employ Mixup operation \cite{zhang2017mixup} in the embedding space to implement this, which can be formulated as:
\begin{equation}
\begin{aligned}
    \tilde{\mathbf{f}}_k^i &=\lambda \mathbf{f}_k^i + (1-\lambda )\mathbf{f}_k^j, \ \ \tilde{\mathbf{y}}^i_k &=\lambda \mathbf{y}^i_k + (1-\lambda ) \mathbf{y}^j_k,
\end{aligned}
\label{FA}
\end{equation}
where $\lambda\in(0,1)$ is a mixing factor. Instead of mixing samples in the input space \cite{2024mixup-cross-modal-retrieval}, our strategy performs the mixing in the embedding space, which is able to mine higher-level information and thus provides additional training signals.
Secondly, since pairwise samples of different modalities are often unavailable and labels contain no modal-specific instance information, we propose a Modality-invariant Representation Recasting mechanism (MRR) to bridge the representation learning on each modality in the common subspace by the learned prior $\mathbf{W}$, thereby obtaining modality-invariant representations while keeping flexibility. Specifically, we first use the prior $\mathbf{W}$ to recast the modality-invariant representation for each semantic category based on the assumption that \emph{the labels are mapped completely accurately}, \ie $|\tilde{\mathbf{f}_k^i} \mathbf{W} - \tilde{\mathbf{y}}^i_k|$ towards 0, which is expressed as:
\begin{equation}
    \begin{aligned}
        |\tilde{\mathbf{f}}_k^i \mathbf{W} - \tilde{\mathbf{y}}^i_k|=0 &\Leftrightarrow \tilde {\mathbf{f}}_k^i \mathbf{W} = \tilde{\mathbf{y}}^i_k \Leftrightarrow\quad\\
              \tilde {\mathbf{f}}_k^i\mathbf{W} \mathbf{W}^{-1}=           \tilde{\mathbf{y}}^i_k\mathbf{W}^{-1} & \Leftrightarrow\quad       \tilde {\mathbf{f}}_k^i= \tilde{\mathbf{y}}^i_k\mathbf{W}^{-1} .
    \end{aligned}
\end{equation}
Thus, the modality-invariant representations ($\tilde{\mathbf{y}}^i_k\mathbf{W}^{-1}$) can be recast by the generalized reversible matrix of $\mathbf{W}$. For convenience, we donate $\mathbf{W}^{-1}$ as $\mathbf{L}$.
Then, we minimize the distance between modality-invariant representation and the representations of corresponding intra-class samples in the common subspace, thus eliminating the cross-modal discrepancy. Technically, the loss for the $k$-th modality is defined as follows: 
\begin{equation}
   \mathcal{J}_{MSE} = \frac{1}{N_k} \sum_{i=1}^{N_k}  \left \| \tilde {\mathbf{f}}_k^i -\tilde{\mathbf{y}}^i_k  \mathbf{L} \right \|_2^2.
   \label{loss: MSE}
\end{equation}
Although $\mathcal{J}_{MSE}$ tries to keep the modality invariance of intra-class samples, it cannot enhance inter-class discriminability between samples well. To this end, we formulate a loss function to enforce similarity among representations of intra-class samples and discriminability among representations of inter-class samples. The loss function is as follows:
\begin{equation}
\begin{split}
     \mathcal{J}_D &= \frac{1}{N_k^2} \sum_{i=1}^{N_k}\sum_{j=1}^{N_k} \left \{ \operatorname{cos}(\tilde{\mathbf{y}}^i_k \mathbf{L},\tilde{\mathbf{y}}^j_k \mathbf{L})-\operatorname{cos}(\tilde {\mathbf{f}}_k^i,\tilde {\mathbf{f}}_k^j) \right \} ^2\\
&+ \frac{1}{N_k^2} \sum_{i=1}^{N_k}\sum_{j=1}^{N_k} \left \{ \operatorname{cos}(\tilde{\mathbf{y}}^i_k \mathbf{L},\tilde {\mathbf{f}}_k^j)-\operatorname{cos}(\tilde {\mathbf{f}}_k^i,\tilde{\mathbf{y}}^j_k \mathbf{L}) \right \} ^2,
\end{split}
\label{loss: D}
\end{equation}
where $\operatorname{cos}(\mathbf{a},\mathbf{b})$ denotes the cosine similarity between $\mathbf{a}$ and $\mathbf{b}$. Clearly, with effective prior learning, minimizing this loss function can robustly enhance the invariance of intra-class sample features in the common subspace, while simultaneously improving the discriminability of inter-class sample features.

Besides, to maintain semantic consistency in the label space between sample representations and ground-truth labels, we exploit the same label loss function as~\Cref{gce_p} in SPL, which is as follows:
\begin{equation}
     \mathcal{J}_L =\frac{1}{qN_k} \sum_{i = 1}^{N_k}\left (  1-\big(\sum_{c=1}^{C} \tilde{y}_{k}^{i,c} \frac{e^{ \tilde {\mathbf{f}}_k^i\mathbf{w}^c}}{ {\textstyle \sum_{j=1}^{C}} e^{ \tilde {\mathbf{f}}_k^i\mathbf{w}^j}} \big) ^q\right ),
     \label{loss: L}
\end{equation}
where $\mathbf{w}^c\in\mathbb{R}^{d\times 1}$ is the $c$-th column vector of $\mathbf{W}$ and $\tilde{y_{k}}^{i,c}$ is the $c$-th element of $\tilde{y}_{k}^{i}$. Finally, we combine the above three loss functions (\Cref{loss: MSE,loss: D,loss: L}) for joint optimization. The total loss function is as follows:
\begin{equation}
    \mathcal{J} = \mathcal{J}_L+ \alpha \mathcal{J}_D + \beta \mathcal{J}_{MSE},
    \label{loss: ALL}
\end{equation}
where $\alpha$ and $\beta$ are the hyperparameters to control the contribution of each component. 
Additionally, the detailed steps of DRCL are shown in Algorithm~\ref{algorithm1} for reproducibility.

\begin{algorithm}
\caption{The training process of our DRCL}
\label{algorithm1}
\KwIn{The dataset $\mathcal{D}=\{ \mathcal{X}_k,\mathcal{Y}_k \}^K_{k=1}$, the encoders $\mathcal{F} = \{F_k(\cdot,\Phi_k)\}^K_k$ of $K$ modalities, and the learning rate $l_r$}
Randomly generate an orthogonal matrix $\mathbf{W}_\text{random}$\;
Randomly initialize $\{\Phi_k\}^K_{k=1}$\;
\For{$k = 1,2,\cdots,K$}{
    $\mathbf{W}_k = \mathbf{W}_\text{random}$\;
    \For{$t = 1,2,\cdots,num\_epochs$}{
        \For{$j=1,2,\cdots,num\_steps$}{
            Sample a mini-batch $(\bar{\mathcal{X}},\bar{\mathcal{Y}})$ from $({\mathcal{X}_k},{\mathcal{Y}_k})$\;
            Calculate the representations for the samples in $\bar{\mathcal{X}}$ by $F_k(\cdot,\Phi_k)$\;
            Calculate $\mathcal{J}_p$ shown in \Cref{gce_p}\;
            Calculate gradients of $\mathbf{W}_k$ and $\Phi_k$ and update parameters by $\mathbf{W}_k =\mathbf{W}_k-l_r\frac{\partial \mathcal{J}_p }{\partial \mathbf{W}_k }$ and $\Phi_k = \Phi_k-l_r\frac{\partial \mathcal{J}_p }{\partial \Phi_k }$\;
        }
    }
    Calculate the quality score $\mathcal{S}_k$ by~\Cref{score}\;
}
Select the weight matrix with the best quality score as the prior matrix $\mathbf{W}$ and reinitialize $\{\Phi_k\}^K_{k=1}$\;
\For{$k = 1,2,\cdots,K$}{
    \For{$t = 1,2,\cdots,num\_epochs$}{
        \For{$j=1,2,\cdots,num\_steps$}{
            Sample a mini-batch $(\bar{\mathcal{X}},\bar{\mathcal{Y}})$ from $({\mathcal{X}_k},{\mathcal{Y}_k})$\;
            Calculate the representations for the samples in $\bar{\mathcal{X}}$ by $F_k(\cdot,\Phi_k)$\;
            Perform FA by \Cref{FA}\;
            Calculate $\mathcal{J}$ shown in \Cref{loss: ALL}\;
            Calculate the gradients of $\Phi_k$ and update parameters by $\Phi_k = \Phi_k-l_r\frac{\partial \mathcal{J}}{\partial \Phi_k }$\;
        }
    }
}
\KwOut{The optimized parameters of $K$ encoders}
\end{algorithm}

\section{Experiments}
To verify the effectiveness of our proposed method, we conducted experiments on five datasets, namely, XMedia~\cite{Xmedia}, XMediaNet \cite{XmediaNet}, Wikipedia \cite{Wiki}, NUS-WIDE \cite{NUS-WIDE}, and INRIA-Websearch \cite{INRIA-Websearch}. First, we compare the proposed DRCL with fifteen state-of-the-art methods to verify its superiority. Subsequently, we perform additional analyses on DRCL, including ablation experiments, hyperparameter selection, the evaluation of different priors, and the assessment of our priors' impact on other advanced methods. Furthermore, the true relevance between samples is measured in terms of their semantic classes, following the approaches in \cite{MRL, SDML, MARS}.

\subsection{Datasets}
The statistics of the five datasets used in our experiments are presented in~\Cref{tab: datasets}. The details of each dataset are provided in the supplementary material.

\begin{table}[]
    \centering
    \caption{General statistics of the five datasets, where “*/*/*” in the “Instance” column indicates the number of samples for training, validation, and testing, respectively. `Websearch' is the abbreviation of the INRIA-Websearch dataset.} 
    \setlength{\tabcolsep}{1pt}
\begin{tabular}{lcrrr}
\toprule[1pt]%
Dataset & Category & Modality & Instance & Feature \\ \hline
\multirow{5}{*}{XMedia} & \multirow{5}{*}{20} & Image & 4,000/500/500 & 4,096d VGG \\
 &  & Text & 4,000/500/500 & 3,000d Bow \\
 &  & Audio clip & 800/100/100 & 29d MFCC \\
 &  & 3D model & 400/50/50 & 4,700d LightField \\
 &  & Video & 969/87/87 & 4,096d C3D \\ \hline
\multirow{5}{*}{XMediaNet} & \multirow{5}{*}{200} & Image & 32,000/4,000/4,000 & 4,096d VGG \\
 &  & Text & 32,000/4,000/4,000 & 300d Doc2Vec \\
 &  & Audio clip & 8,000/1,000/1,000 & 672d MFCC \\
 &  & 3D model & 1,600/200/200 & 4,700d LightField \\
 &  & Video & 8,000/1,000/1,000 & 4,096d C3D \\ \hline
\multirow{2}{*}{Wikipedia} & \multirow{2}{*}{10} & Image & 2,173/231/462 & 4,096d VGG \\
 &  & Text & 2,173/231/462 & 300d Doc2Vec \\ \hline
\multirow{2}{*}{NUS-WIDE} & \multirow{2}{*}{10} & Image & 8,000/1,000/1,000 & 4,096d VGG \\
 &  & Text & 8,000/1,000/1,000 & 300d Doc2Vec \\ \hline
\multirow{2}{*}{Websearch} & \multirow{2}{*}{100} & Image & 9,000/1,332/4,366 & 4,096d VGG \\
 &  & Text & 9,000/1,332/4,366 & 1,000d LDA \\ \bottomrule[1pt]%
\end{tabular}
\label{tab: datasets}
\end{table}

\begin{table*}[]
    \centering
    \caption{Comparison of MAP@all scores on the XMedia and XMediaNet datasets. The best and second-best results are in \textbf{bold} and \underline{underlined}, respectively. The methods requiring repeated sampling are marked with `*'.}
  \setlength\tabcolsep{1pt}
\begin{tabular}{c|c|ccccccccccccccccccccc}
\toprule[1pt]%
\multirow{2}{*}{Dataset}&Methods & \multicolumn{4}{c|}{Image} & \multicolumn{4}{c|}{Text} & \multicolumn{4}{c|}{Audio} & \multicolumn{4}{c|}{3D} & \multicolumn{4}{c|}{Video} & \multirow{2}{*}{Avg} \\ \cline{2-22}
&Target & Text & Audio & 3D & \multicolumn{1}{c|}{Video} & Image & Audio & 3D & \multicolumn{1}{c|}{Video} & Image & Text & 3D & \multicolumn{1}{c|}{Video} & Image & Text & Audio & \multicolumn{1}{c|}{Video} & Image & Text & Audio & \multicolumn{1}{c|}{3D} &  \\ \hline
\multirow{10}{*}{\rotatebox{90}{XMedia}} & MCCA& 0.752 & 0.099 & 0.120 & \multicolumn{1}{c|}{0.126} & 0.773 & 0.106 & 0.115 & \multicolumn{1}{c|}{0.086} & 0.067 & 0.074 & 0.121 & \multicolumn{1}{c|}{0.099} & 0.096 & 0.057 & 0.125 & \multicolumn{1}{c|}{0.101} & 0.090 & 0.088 & 0.105 & \multicolumn{1}{c|}{0.121} & 0.166 \\
&SDML& \underline{0.901} & 0.480 & \underline{0.696} & \multicolumn{1}{c|}{\underline{0.658}
} & \textbf{0.917} & 0.501 & \underline{0.724} & \multicolumn{1}{c|}{\underline{0.687}} & 0.474 & 0.489 & 0.436 & \multicolumn{1}{c|}{0.373} & \underline{0.682} & \underline{0.699} & 0.383 & \multicolumn{1}{c|}{\textbf{0.552}} & 0.585 & 0.597 & 0.327 & \multicolumn{1}{c|}{\underline{0.516}} & 0.584 \\
&MAN*&0.871 & \underline{0.581} & 0.556 & \multicolumn{1}{c|}{0.631} & 0.893 & \underline{0.617} & 0.593 & \multicolumn{1}{c|}{0.669} & \underline{0.568} & \underline{0.596} & \underline{0.460} & \multicolumn{1}{c|}{\underline{0.439}} & 0.560 & 0.574 & 0.370 & \multicolumn{1}{c|}{0.457} & \underline{0.588} & \underline{0.614} & \underline{0.404} & \multicolumn{1}{c|}{0.439} & 0.574 \\
&MRL* & 0.795 & 0.309 & 0.406 & \multicolumn{1}{c|}{0.423} & 0.776 & 0.290 & 0.405 & \multicolumn{1}{c|}{0.396} & 0.292 & 0.296 & 0.170 & \multicolumn{1}{c|}{0.145} & 0.419 & 0.416 & 0.162 & \multicolumn{1}{c|}{0.333} & 0.549 & 0.511 & 0.134 & \multicolumn{1}{c|}{0.342} & 0.379 \\
&ELRCMR*& 0.814 & 0.329 & 0.419 & \multicolumn{1}{c|}{0.370} & 0.779 & 0.312 & 0.400 & \multicolumn{1}{c|}{0.320} & 0.289 & 0.271 & 0.138 & \multicolumn{1}{c|}{0.128} & 0.385 & 0.399 & 0.182 & \multicolumn{1}{c|}{0.303} & 0.493 & 0.491 & 0.128 & \multicolumn{1}{c|}{0.386} & 0.367 \\
&MARS & 0.887 & 0.569 & 0.643 & \multicolumn{1}{c|}{0.642} & 0.888 & 0.593 & 0.667 & \multicolumn{1}{c|}{0.671} & 0.534 & 0.565 & 0.454 & \multicolumn{1}{c|}{0.409} & 0.652 & 0.667 & \underline{0.430} & \multicolumn{1}{c|}{0.488} & 0.584 & 0.605 & 0.381 & \multicolumn{1}{c|}{0.473} & \underline{0.590} \\
&GNN4CMR* & 0.875 & 0.453 & 0.624 & \multicolumn{1}{c|}{0.627} & 0.880 & 0.472 & 0.642 & \multicolumn{1}{c|}{0.617} & 0.475 & 0.480 & 0.415 & \multicolumn{1}{c|}{0.244} & 0.595 & 0.587 & 0.306 & \multicolumn{1}{c|}{0.446} & 0.562 & 0.563 & 0.238 & \multicolumn{1}{c|}{0.459} & 0.528 \\
&RONO* & 0.896 & 0.469 & 0.656 & \multicolumn{1}{c|}{0.576} & 0.906 & 0.442 & 0.665 & \multicolumn{1}{c|}{0.637} & 0.474 & 0.480 & 0.124 & \multicolumn{1}{c|}{0.124} & 0.584 & 0.619 & 0.199 & \multicolumn{1}{c|}{0.413} & 0.555 & 0.589 & 0.141 & \multicolumn{1}{c|}{0.496} & 0.502 \\
&COXI* & 0.896 & 0.225 & 0.206 & \multicolumn{1}{c|}{0.342} & \underline{0.910} & 0.222 & 0.271 & \multicolumn{1}{c|}{0.283} & 0.206 & 0.218 & 0.106 & \multicolumn{1}{c|}{0.153} & 0.156 & 0.163 & 0.102 & \multicolumn{1}{c|}{0.126} & 0.316 & 0.298 & 0.131 & \multicolumn{1}{c|}{0.169} & 0.275 \\
&DRCL& \textbf{0.902} & \textbf{0.624} & \textbf{0.698} & \multicolumn{1}{c|}{\textbf{0.662}} & 0.909 & \textbf{0.666} & \textbf{0.729} & \multicolumn{1}{c|}{\textbf{0.695}} & \textbf{0.593} & \textbf{0.625} & \textbf{0.506} & \multicolumn{1}{c|}{\textbf{0.443}} & \textbf{0.688} & \textbf{0.712} & \textbf{0.460} & \multicolumn{1}{c|}{\underline{0.522}} & \textbf{0.594} & \textbf{0.621} & \textbf{0.421} & \multicolumn{1}{c|}{\textbf{0.539}} & \textbf{0.630} \\ \bottomrule[1pt]%
\multirow{10}{*}{\rotatebox{90}{XMediaNet}}&MCCA & 0.090 & 0.012 & 0.021 & \multicolumn{1}{c|}{0.011} & 0.098 & 0.012 & 0.022 & \multicolumn{1}{c|}{0.011} & 0.008 & 0.007 & 0.023 & \multicolumn{1}{c|}{0.013} & 0.006 & 0.006 & 0.011 & \multicolumn{1}{c|}{0.008} & 0.007 & 0.007 & 0.012 & \multicolumn{1}{c|}{0.018} & 0.020 \\
&SDML& 0.561 & 0.351 & 0.293 & \multicolumn{1}{c|}{0.357} & 0.576 & 0.289 & 0.240 & \multicolumn{1}{c|}{0.294} & \underline{0.369} & 0.293 & 0.146 & \multicolumn{1}{c|}{0.191} & 0.345 & 0.269 & 0.165 & \multicolumn{1}{c|}{0.190} & 0.358 & 0.290 & 0.189 & \multicolumn{1}{c|}{0.165} & 0.297 \\
&MAN*& 0.007 & 0.013 & 0.020 & \multicolumn{1}{c|}{0.012} & 0.010 & 0.013 & 0.021 & \multicolumn{1}{c|}{0.015} & 0.008 & 0.007 & 0.028 & \multicolumn{1}{c|}{0.014} & 0.007 & 0.007 & 0.016 & \multicolumn{1}{c|}{0.011} & 0.008 & 0.007 & 0.014 & \multicolumn{1}{c|}{0.030} & 0.013 \\
&MRL*& 0.084 & 0.027 & 0.025 & \multicolumn{1}{c|}{0.025} & 0.106 & 0.028 & 0.024 & \multicolumn{1}{c|}{0.025} & 0.014 & 0.011 & 0.023 & \multicolumn{1}{c|}{0.014} & 0.014 & 0.011 & 0.016 & \multicolumn{1}{c|}{0.017} & 0.016 & 0.015 & 0.019 & \multicolumn{1}{c|}{0.022} & 0.027 \\
&ELRCMR*& 0.056 & 0.024 & 0.017 & \multicolumn{1}{c|}{0.024} & 0.073 & 0.024 & 0.018 & \multicolumn{1}{c|}{0.023} & 0.017 & 0.011 & 0.014 & \multicolumn{1}{c|}{0.013} & 0.010 & 0.009 & 0.018 & \multicolumn{1}{c|}{0.037} & 0.017 & 0.015 & 0.016 & \multicolumn{1}{c|}{0.031} & 0.023 \\
&MARS& 0.584 & \underline{0.356} & \underline{0.310} & \multicolumn{1}{c|}{0.368} & 0.574 & \underline{0.309} & \underline{0.263} & \multicolumn{1}{c|}{\underline{0.320}} & 0.356 & \underline{0.311} & \underline{0.170} & \multicolumn{1}{c|}{\underline{0.197}} & \underline{0.346} & \underline{0.289}  & \underline{0.191} & \multicolumn{1}{c|}{\underline{0.220}} & \underline{0.374} & \underline{0.327} & \underline{0.201} & \multicolumn{1}{c|}{\textbf{0.198}} & \underline{0.313} \\
&GNN4CMR*& 0.577 & 0.292 & 0.266 & \multicolumn{1}{c|}{\underline{0.384}} & 0.582 & 0.217 & 0.232 & \multicolumn{1}{c|}{0.312} & 0.288 & 0.219 & 0.126 & \multicolumn{1}{c|}{0.015} & 0.307 & 0.233 & 0.115 & \multicolumn{1}{c|}{0.163} & \underline{0.374} & 0.303 & 0.014 & \multicolumn{1}{c|}{\underline{0.167}} & 0.259 \\
&RONO*& 0.173 & 0.158 & 0.244 & \multicolumn{1}{c|}{0.296} & 0.144 & 0.034 & 0.092 & \multicolumn{1}{c|}{0.057} & 0.151 & 0.034 & 0.066 & \multicolumn{1}{c|}{0.011} & 0.213 & 0.073 & 0.058 & \multicolumn{1}{c|}{0.127} & 0.283 & 0.068 & 0.012 & \multicolumn{1}{c|}{0.114} & 0.120 \\
&COXI*& \textbf{0.645} & 0.066 & 0.042 & \multicolumn{1}{c|}{0.139} & \underline{0.634} & 0.065 & 0.050 & \multicolumn{1}{c|}{0.137} & 0.087 & 0.077 & 0.035 & \multicolumn{1}{c|}{0.012} & 0.027 & 0.024 & 0.028 & \multicolumn{1}{c|}{0.021} & 0.153 & 0.130 & 0.013 & \multicolumn{1}{c|}{0.027} & 0.121 \\
&DRCL & \underline{0.640} & \textbf{0.408} & \textbf{0.313} & \multicolumn{1}{c|}{\textbf{0.415}} & \textbf{0.642} & \textbf{0.358} & \textbf{0.274} & \multicolumn{1}{c|}{\textbf{0.364}} & \textbf{0.412} & \textbf{0.356} & \textbf{0.186} & \multicolumn{1}{c|}{\textbf{0.236}} & \textbf{0.362} & \textbf{0.310} & \textbf{0.209} & \multicolumn{1}{c|}{\textbf{0.236}} & \textbf{0.409} & \textbf{0.343} & \textbf{0.232} & \multicolumn{1}{c|}{\textbf{0.198}} & \textbf{0.345 }\\ \bottomrule[1pt]%
\end{tabular}
    \label{all-xmedia}
\end{table*}

\begin{table}[]
    \centering
    \setlength\tabcolsep{2.0pt}
    \caption{Comparison of MAP@all scores on the Wikipedia, NUS-WIDE, and INRIA-Websearch datasets. The best and second-best results are in \textbf{bold} and \underline{underlined}, respectively. The `I', and `T' abbreviations stand for Image and Text, respectively.}
\begin{tabular}{c|ccc|ccc|ccc}
\toprule[1pt]%
\multirow{2}{*}{Methods} & \multicolumn{3}{c|}{Wikipedia} & \multicolumn{3}{c|}{INRIA-Websearch} & \multicolumn{3}{c}{NUS-WIDE} \\ \cline{2-10} 
 & \multicolumn{1}{c|}{I →T} & \multicolumn{1}{c|}{T→I} & Avg & \multicolumn{1}{c|}{I →T} & \multicolumn{1}{c|}{T→I} & Avg & \multicolumn{1}{c|}{I →T} & \multicolumn{1}{c|}{T→I} & Avg \\ \hline
MCCA & \multicolumn{1}{c|}{0.296} & \multicolumn{1}{c|}{0.283} & 0.290 & \multicolumn{1}{c|}{0.369} & \multicolumn{1}{c|}{0.393} & 0.381 & \multicolumn{1}{c|}{0.387} & \multicolumn{1}{c|}{0.392} & 0.390 \\
CMCP & \multicolumn{1}{c|}{0.524} & \multicolumn{1}{c|}{0.489} & 0.507 & \multicolumn{1}{c|}{0.343} & \multicolumn{1}{c|}{0.359} & 0.351 & \multicolumn{1}{c|}{0.551} & \multicolumn{1}{c|}{0.572} & 0.562 \\
ACMR & \multicolumn{1}{c|}{0.434} & \multicolumn{1}{c|}{0.414} & 0.424 & \multicolumn{1}{c|}{0.420} & \multicolumn{1}{c|}{0.426} & 0.423 & \multicolumn{1}{c|}{0.537} & \multicolumn{1}{c|}{0.542} & 0.540 \\
MMSAE & \multicolumn{1}{c|}{0.413} & \multicolumn{1}{c|}{0.383} & 0.398 & \multicolumn{1}{c|}{0.144} & \multicolumn{1}{c|}{0.141} & 0.143 & \multicolumn{1}{c|}{0.531} & \multicolumn{1}{c|}{0.549} & 0.540 \\
DSCMR & \multicolumn{1}{c|}{0.489} & \multicolumn{1}{c|}{0.462} & 0.476 & \multicolumn{1}{c|}{0.538} & \multicolumn{1}{c|}{\underline{0.572}} & 0.555 & \multicolumn{1}{c|}{0.578} & \multicolumn{1}{c|}{0.587} & 0.583 \\
SDML & \multicolumn{1}{c|}{0.546} & \multicolumn{1}{c|}{0.503} & 0.525 & \multicolumn{1}{c|}{0.542} & \multicolumn{1}{c|}{0.571} & 0.557 & \multicolumn{1}{c|}{\underline{0.597}} & \multicolumn{1}{c|}{\underline{0.592}} & \underline{0.595} \\
MAN & \multicolumn{1}{c|}{0.449} & \multicolumn{1}{c|}{0.421} & 0.435 & \multicolumn{1}{c|}{0.119} & \multicolumn{1}{c|}{0.126} & 0.123 & \multicolumn{1}{c|}{0.484} & \multicolumn{1}{c|}{0.528} & 0.506 \\
DRSL & \multicolumn{1}{c|}{0.498} & \multicolumn{1}{c|}{0.472} & 0.485 & \multicolumn{1}{c|}{0.479} & \multicolumn{1}{c|}{0.509} & 0.494 & \multicolumn{1}{c|}{0.517} & \multicolumn{1}{c|}{0.555} & 0.536 \\
MRL & \multicolumn{1}{c|}{\underline{0.549}} & \multicolumn{1}{c|}{\underline{0.512}} & \underline{0.531} & \multicolumn{1}{c|}{0.313} & \multicolumn{1}{c|}{0.304} & 0.309 & \multicolumn{1}{c|}{0.572} & \multicolumn{1}{c|}{0.581} & 0.577 \\
ALGCN & \multicolumn{1}{c|}{0.485} & \multicolumn{1}{c|}{0.451} & 0.468 & \multicolumn{1}{c|}{0.415} & \multicolumn{1}{c|}{0.411} & 0.413 & \multicolumn{1}{c|}{0.569} & \multicolumn{1}{c|}{0.570} & 0.570 \\
ELRCMR & \multicolumn{1}{c|}{0.543} & \multicolumn{1}{c|}{0.501} & 0.522 & \multicolumn{1}{c|}{0.300} & \multicolumn{1}{c|}{0.283} & 0.292 & \multicolumn{1}{c|}{0.549} & \multicolumn{1}{c|}{0.567} & 0.558 \\
MARS & \multicolumn{1}{c|}{0.531} & \multicolumn{1}{c|}{0.479} & 0.505 & \multicolumn{1}{c|}{0.545} & \multicolumn{1}{c|}{0.564} & 0.555 & \multicolumn{1}{c|}{0.551} & \multicolumn{1}{c|}{0.555} & 0.553 \\
GNN4CMR & \multicolumn{1}{c|}{0.521} & \multicolumn{1}{c|}{0.480} & 0.501 & \multicolumn{1}{c|}{0.523} & \multicolumn{1}{c|}{0.538} & 0.531 & \multicolumn{1}{c|}{0.595} & \multicolumn{1}{c|}{0.591} & 0.593 \\
RONO & \multicolumn{1}{c|}{0.521} & \multicolumn{1}{c|}{0.473} & 0.497 & \multicolumn{1}{c|}{0.470} & \multicolumn{1}{c|}{0.478} & 0.474 & \multicolumn{1}{c|}{0.559} & \multicolumn{1}{c|}{0.590} & 0.575 \\
COXI & \multicolumn{1}{c|}{0.547} & \multicolumn{1}{c|}{0.499} & 0.523 & \multicolumn{1}{c|}{\underline{0.553}} & \multicolumn{1}{c|}{0.571} & \underline{0.562} & \multicolumn{1}{c|}{0.571} & \multicolumn{1}{c|}{0.576} & 0.574 \\
DRCL & \multicolumn{1}{c|}{\textbf{0.562}} & \multicolumn{1}{c|}{\textbf{0.537}} & \textbf{0.550} & \multicolumn{1}{c|}{\textbf{0.567}} & \multicolumn{1}{c|}{\textbf{0.588}} & \textbf{0.578} & \multicolumn{1}{c|}{\textbf{0.606}} & \multicolumn{1}{c|}{\textbf{0.615}} & \textbf{0.611} \\ \bottomrule[1pt]%
\end{tabular}
\label{all-Two-modalities}
\end{table}

\subsection{Implementation Details}
In DRCL, each subnetwork has independent parameters but shares the same network structure that comprises two fully connected layers and a representation layer. The first two layers consist of 4096 neurons each, employing the ReLU activation function. For the XMedia, XMediaNet, and INRIA-Websearch datasets, the number of neural units in the representation layer is set to 512, while for the Wikipedia and NUS-WIDE datasets, the number is 2048. Note that the output of the representation layer is normalized. For all datasets, we set the hyperparameters as $\alpha=0.1$ and $\beta=0.1$ based on our parameter sensitivity analysis in~\Cref{sec_hya} and the mixing factor $\lambda$ as $0.9$. All experiments are conducted on a single GeForce RTX3090Ti 24GB GPU.  

\subsection{Evaluation Metric}
To assess the efficacy of various methodologies, we employ mean average precision (MAP) as the evaluation metric. The MAP score for the top $N_{rank}$ rankings can be formulated as:
\begin{equation}
    MAP = \frac{1}{N_t} \sum_{i=1}^{N_t} \frac{1}{R_i} \sum_{k=1}^{N_{rank}} \frac{R_{i,k}}{k} \times R_{i,flag}
    \label{eq_map}
\end{equation}
where $N_t$ is the number of samples in the test set, $R_i$ is the number of samples relevant to the $i$-th query in the top-$N_{rank}$ returned result, and $R_{i,k}$ is the number of samples relevant to the $i$-th query in the top-$k$ returned result. Note that $R_{i,flag}$ = 1 if the $k$-th returned result is relevant to the $i$-th query, otherwise $R_{i,flag}$ = 0. In our experiments, we exploit MAP scores for both overall rankings (MAP@all) and the top 50 rankings (MAP@50) as retrieval metrics to evaluate retrieval performance. The results of MAP@50 scores are provided in the supplementary material.

\subsection{Compared Methods}
To validate the effectiveness of our proposed approach, we conduct a comparative analysis against fifteen state-of-the-art methodologies, namely, MCCA~(TIP’02)~\cite{MCCA_set}, CMCP~(ICASSP'12)~\cite{CMCP}, ACMR~(ACM MM'17)~\cite{ACMR}, MMSAE~(IJON'19)~\cite{mmsae}, DSCMR~(CVPR'19)~\cite{DSCMR}, SDML~(SIGIR'19)~\cite{SDML}, MAN~(KBS'19)~\cite{MAN}, DRSL~(INS'21)~\cite{DRSL}, MRL~(CVPR'21)~\cite{MRL}, ALGCN~(TMM'21)~\cite{ALGCN}, ELRCMR~(ACM MM'22)~\cite{ELRCMR}, MARS~(TCSVT'22)~\cite{MARS}, GNN4CMR~(TPAMI'23)~\cite{GNN4CMR}, RONO~(CVPR'23)~\cite{RONO}, and COXI~(IJMIR'24)~\cite{COXI}. Since RONO~\cite{RONO} aims at cross-modal retrieval between images and 3D point clouds, we replace the backbones used in~\cite{DSCMR} with the original backbones. For the remaining methods, we utilize their original backbones.  
These baselines can be divided into two groups: 1) Paired-oriented methods that require paired modal data. 2) Paired-free methods do not require paired modal data.
Among these baselines, CMCP, ACMR, MMSAE, DSCMR, DRSL, ALGCN, GNN4CMR, RONO, and COXI are the paired-oriented methods that can only handle two modalities simultaneously. In our experiments, for these baselines except for GNN4CMR, RONO, and COXI, we only conducted experiments on the Wikipedia, NUS-WIDE, and INRIA-Websearch datasets. For GNN4CMR, RONO, and COXI, the experiments encompassed not only the Wikipedia, NUS-WIDE, and Websearch datasets but also extended to the XMedia and XMediaNet datasets by performing multiple experiments in a pairwise manner on every two modalities. By contrast, MAN, MRL, and ELRCMR are paired-oriented methods that can only handle multiple modalities simultaneously. For all paired-oriented methods, we employ repeated sampling to ensure cross-modal data in pairs during training. As for paired-free methods, MCCA, SDML, MARS, and our DRCL, we conduct experiments directly without any processing of the datasets. Besides, \emph{note that all results are reproduction results on the same dataset settings}.

\subsection{Comparison with the State-of-the-Art Methods}
As shown in~\Cref{all-xmedia}, compared with nine state-of-the-art methods, the average MAP@all of our DRCL on the XMedia dataset surpasses the second-best baseline MARS by 4.0\%. For the XMediaNet dataset, our DRCL achieves average MAPs that are 3.2\% higher than that of the second-best baseline MARS. Additionally, MCCA exhibits poor performance due to the failure to leverage semantic information in labels and reliance on linear modeling. 
Notably, the large number of categories (200) in the XMediaNet dataset increases the learning difficulty, resulting in a significant performance decrease across all methods. Some paired-oriented methods, such as MAN, MRL, ELRCMR, GNN4CMR, RONO, and COXI, directly push representations of true sample pairs close to eliminate the cross-modal gap. However, for the XMedia and XMediaNet datasets, only images and texts are paired while other modalities can only construct sample pairs through repeated sampling. As a result, the representations of intra-class samples learned from them are not compact enough, resulting in lower performance. In contrast, our DRCL focuses on the consistency of intra-class samples, thus achieving good performance. Although SDML and MARS also focus on the consistency of intra-class samples like our DRCL, DRCL outperforms them as we learned the best prior which liberates constraints on the semantic alignment between sample representations and ground-truth semantic labels. The comprehensive experimental results in these tables underscore the effectiveness of our DRCL.

\begin{figure*}[t]
    \centering
    \captionsetup[subfloat]{labelsep=none,format=plain}
    \subfloat[Wikipedia(I2T)]{
		\includegraphics[scale=0.315]{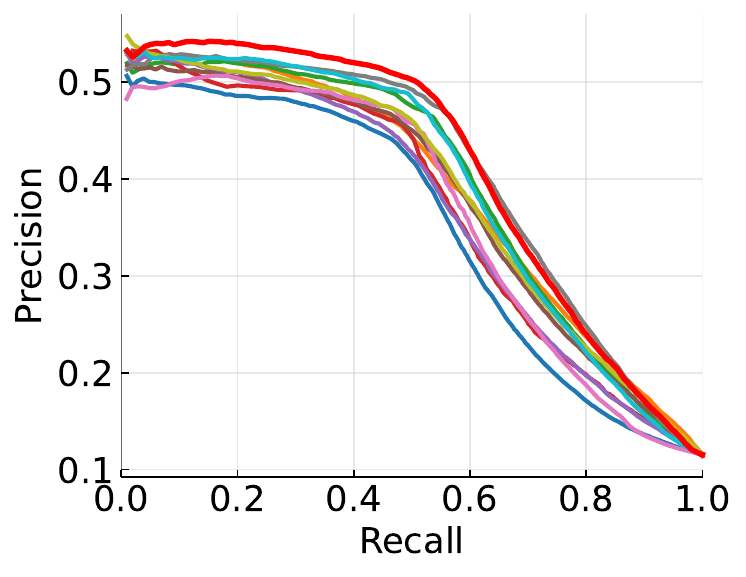}}
    \subfloat[Wikipedia(T2I)]{
		\includegraphics[scale=0.315]{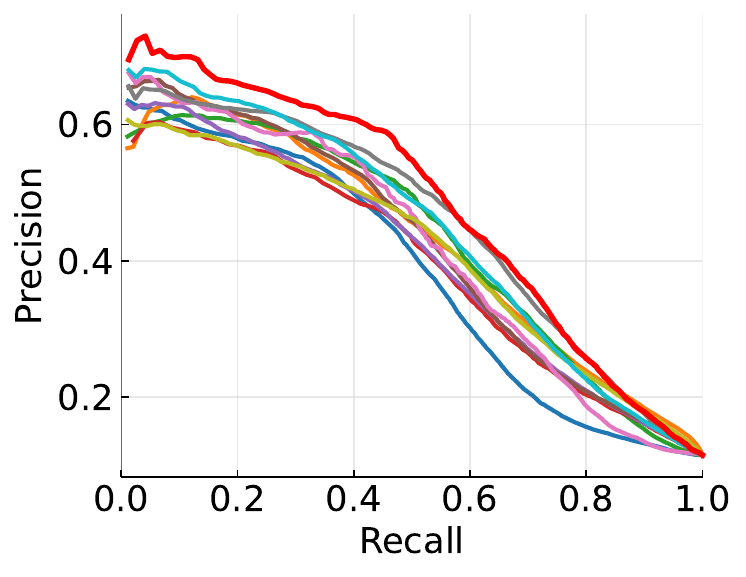}}
    \subfloat[NUS-WIDE(I2T)]{
		\includegraphics[scale=0.315]{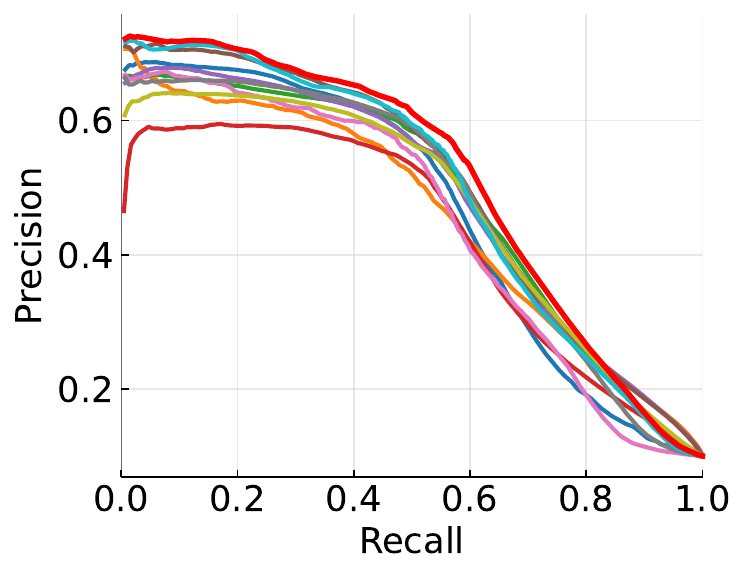}}
  \subfloat[NUS-WIDE(T2I)]{
		\includegraphics[scale=0.315]{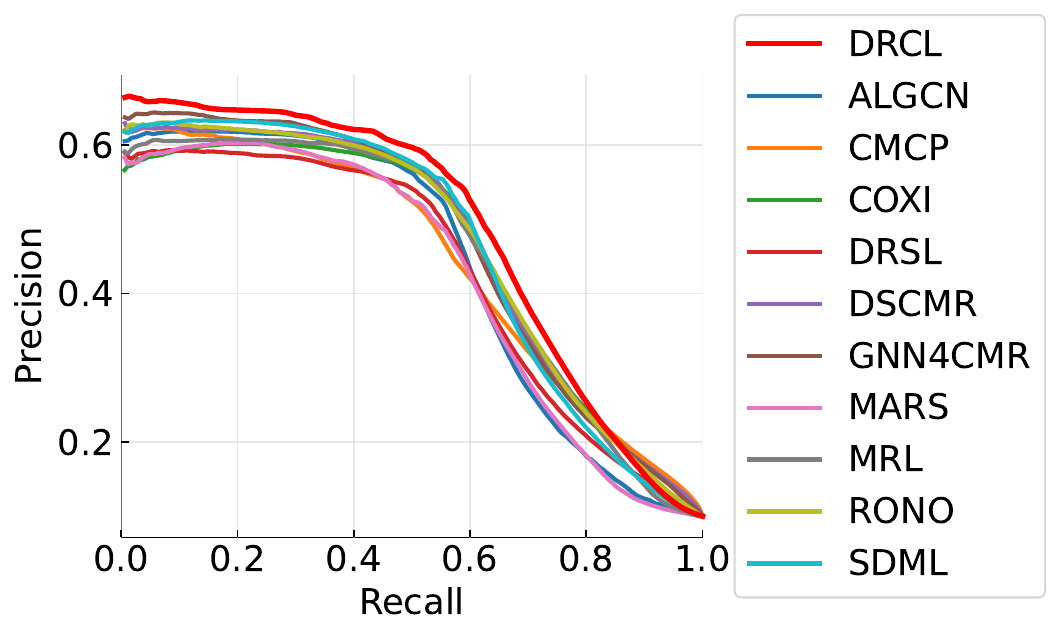}}
  \\
   \captionsetup{labelsep=colon}
  \caption{Precision-recall curves on the Wikipedia and NUS-WIDE datasets. See the supplementary material for more results.}
  \label{fig: PR} 
\end{figure*}

From~\Cref{all-Two-modalities}, it can be observed that DRCL obtains the highest average MAP@all scores on the Wikipedia, INRIA-Websearch, and NUS-WIDE datasets, which are 0.550, 0.578, and 0.611, respectively, while the second-best scores are 0.531, 0.562, and 0.595, respectively. 
It is noteworthy that Wikipedia, INRIA-Websearch, and NUS-WIDE are the datasets that contain only two modal data in pairs. Hence, most methods that require paired modal data have decent performance on these datasets. However, the INRIA-Websearch dataset has more categories, increasing the learning complexity, some methods perform poorly on this dataset, \eg MMSAE and MAN. Besides, our DRCL also can outperform SDML and MARS on the INRIA-Websearch dataset thanks to that we learned the best prior. Overall, our DRCL shows excellent performance on all datasets, which is enough to prove the stability and superiority of our method. Also, to further evaluate the performance of the proposed DRCL, precision-recall curves in terms of two cross-modal retrieval tasks (\ie I2T and T2I) on the Wikipedia and NUS-WIDE datasets are plotted in~\Cref{fig: PR}. In the figure, recall denotes the proportion of samples that the model correctly retrieves as relevant among all samples that are actually relevant to the query, while precision denotes the proportion of samples that are actually relevant among all samples predicted by the model to be relevant to the query. From the figure, it can be observed that our DRCL has higher precision at the same recall rate, demonstrating the superior performance of the proposed DRCL, which has a consistent conclusion with MAP scores.

\begin{table}[]
    \centering
    \caption{Ablation experiment settings, where `*' in the column of  `$q$' indicates the same settings as DRCL.  $\mathbf{W^{-1}}$, $\mathbf{W^\top}$, and \underline{FA} represent the inverse matrix of the prior matrix, the transpose matrix of the prior matrix, and conducting FA in the input space, respectively. `$\checkmark$' represents used.} 
    \begin{tabular}{l|c|cccccc}
        \toprule[1pt]%
        \multirow{2}{*}{No.} & \multicolumn{1}{c|}{\multirow{2}{*}{SPL}} & \multicolumn{6}{c}{RSC} \\ 
         & \multicolumn{1}{c|}{} & $\mathcal{J}_L$ & $\mathcal{J}_D$ & $\mathcal{J}_{MSE}$ & $q$ & $\mathbf{L}$ & FA \\ \hline
    \#1 &  & $\checkmark$ & $\checkmark$ & $\checkmark$ & * & $\mathbf{W^{-1}}$& $\checkmark$ \\
    \#2 & $\checkmark$ &  & $\checkmark$ & $\checkmark$ & * & $\mathbf{W^{-1}}$& $\checkmark$\\
    \#3 & $\checkmark$ & $\checkmark$ &  &$\checkmark$ & * & $\mathbf{W^{-1}}$& $\checkmark$ \\
    \#4 & $\checkmark$ & $\checkmark$ & $\checkmark$ &  & * & $\mathbf{W^{-1}}$& $\checkmark$ \\
    \#5 & $\checkmark$ & $\checkmark$ & $\checkmark$ & $\checkmark$ & 0.01 & $\mathbf{W^{-1}}$& $\checkmark$ \\
    \#6 & $\checkmark$ & $\checkmark$ & $\checkmark$ & $\checkmark$ & 0.50 & $\mathbf{W^{-1}}$& $\checkmark$ \\
    \#7 & $\checkmark$ & $\checkmark$ & $\checkmark$ & $\checkmark$ & 1.00 & $\mathbf{W^{-1}}$& $\checkmark$ \\
    \#8 & $\checkmark$ & $\checkmark$ & $\checkmark$ & $\checkmark$ & 2.00 & $\mathbf{W^{-1}}$& $\checkmark$ \\
    \#9 & $\checkmark$ & $\checkmark$ & $\checkmark$ & $\checkmark$ & * & $\mathbf{W^\top}$& $\checkmark$ \\
    \#10 & $\checkmark$ & $\checkmark$ & $\checkmark$& $\checkmark$ & * & $\mathbf{W^{-1}}$&  \\
    \#11 & $\checkmark$ & $\checkmark$ & $\checkmark$& $\checkmark$ & * & $\mathbf{W^{-1}}$&  \underline{FA} \\
    Ours & $\checkmark$ & $\checkmark$ & $\checkmark$ & $\checkmark$ & * & $\mathbf{W^{-1}}$& $\checkmark$ \\ \bottomrule[1pt]%
    \end{tabular}
    \label{tab: Ablation study setting}
\end{table}

\begin{table*}[]
    \centering
    \caption{Performance (MAP@all) on the XMedia dataset with different ablation settings. Best scores are in \textbf{Bold}.}
     \setlength\tabcolsep{2pt}
    \begin{tabular}{c|ccccccccccccccccccccc}
\toprule[1pt]%
 & \multicolumn{21}{c}{XMedia} \\ \cline{2-22} 
 & \multicolumn{4}{c|}{Image} & \multicolumn{4}{c|}{Text} & \multicolumn{4}{c|}{Audio} & \multicolumn{4}{c|}{3D} & \multicolumn{4}{c|}{Video} & \multirow{2}{*}{Avg} \\ \cline{2-21}
 No.& Text & Audio & 3D & \multicolumn{1}{c|}{Video} & Image & Audio & 3D & \multicolumn{1}{c|}{Video} & Image & Text & 3D & \multicolumn{1}{c|}{Video} & Image & Text & Audio & \multicolumn{1}{c|}{Video} & Image & Text & Audio & \multicolumn{1}{c|}{3D} &  \\ \hline
\#1 & 0.896 & 0.607 & 0.644 & \multicolumn{1}{c|}{0.659} & 0.902 & 0.642 & 0.671 & \multicolumn{1}{c|}{0.690} & 0.578 & 0.612 & 0.490 & \multicolumn{1}{c|}{\textbf{0.461}} & 0.634 & 0.661 & 0.444 & \multicolumn{1}{c|}{0.485} & 0.586 & 0.620 & 0.416 & \multicolumn{1}{c|}{0.507} & 0.610 \\
\#2 & 0.073 & 0.082 & 0.098 & \multicolumn{1}{c|}{0.137} & 0.049 & 0.119 & 0.162 & \multicolumn{1}{c|}{0.111} & 0.082 & 0.104 & 0.125 & \multicolumn{1}{c|}{0.086} & 0.068 & 0.115 & 0.116 & \multicolumn{1}{c|}{0.110} & 0.127 & 0.127 & 0.083 & \multicolumn{1}{c|}{0.115} & 0.104 \\
\#3 & 0.900 & 0.610 & 0.649 & \multicolumn{1}{c|}{\textbf{0.663}} & 0.905 & 0.647 & 0.685 & \multicolumn{1}{c|}{0.691} & 0.588 & \textbf{0.625} & 0.469 & \multicolumn{1}{c|}{0.420} & 0.642 & 0.667 & 0.452 & \multicolumn{1}{c|}{0.496} & 0.580 & 0.622 & 0.417 & \multicolumn{1}{c|}{0.480} & 0.610 \\
\#4 & 0.901 & \textbf{0.624} & 0.638 & \multicolumn{1}{c|}{\textbf{0.663}} & 0.904 & 0.660 & 0.661 & \multicolumn{1}{c|}{\textbf{0.695}} & 0.589 & \textbf{0.625} & 0.482 & \multicolumn{1}{c|}{0.438} & 0.683 & 0.704 & 0.466 & \multicolumn{1}{c|}{\textbf{0.525}} & 0.585 & 0.618 & 0.416 & \multicolumn{1}{c|}{0.515} & 0.620 \\
\#5 & 0.898 & 0.561 & 0.633 & \multicolumn{1}{c|}{0.637} & \textbf{0.909} & 0.603 & 0.669 & \multicolumn{1}{c|}{0.669} & 0.534 & 0.569 & 0.455 & \multicolumn{1}{c|}{0.375} & 0.603 & 0.627 & 0.359 & \multicolumn{1}{c|}{0.443} & 0.588 & \textbf{0.625} & 0.367 & \multicolumn{1}{c|}{0.490} & 0.581 \\
\#6 & 0.899 & 0.609 & 0.570 & \multicolumn{1}{c|}{0.629} & 0.898 & 0.650 & 0.594 & \multicolumn{1}{c|}{0.656} & 0.584 & 0.623 & 0.450 & \multicolumn{1}{c|}{0.431} & 0.545 & 0.564 & 0.384 & \multicolumn{1}{c|}{0.440} & 0.574 & 0.616 & \textbf{0.426} & \multicolumn{1}{c|}{0.459} & 0.580 \\
\#7 & 0.892 & 0.545 & 0.650 & \multicolumn{1}{c|}{0.591} & 0.887 & 0.579 & 0.676 & \multicolumn{1}{c|}{0.625} & 0.541 & 0.585 & 0.459 & \multicolumn{1}{c|}{0.387} & 0.554 & 0.585 & 0.362 & \multicolumn{1}{c|}{0.419} & 0.557 & 0.595 & 0.374 & \multicolumn{1}{c|}{0.459} & 0.566 \\
\#8 & 0.813 & 0.455 & 0.487 & \multicolumn{1}{c|}{0.553} & 0.787 & 0.443 & 0.415 & \multicolumn{1}{c|}{0.555} & 0.360 & 0.371 & 0.268 & \multicolumn{1}{c|}{0.271} & 0.382 & 0.375 & 0.190 & \multicolumn{1}{c|}{0.338} & 0.521 & 0.546 & 0.265 & \multicolumn{1}{c|}{0.406} & 0.440 \\
\#9 & 0.898 & 0.592 & 0.681 & \multicolumn{1}{c|}{0.635} & 0.903 & 0.624 & 0.714 & \multicolumn{1}{c|}{0.662} & 0.567 & 0.600 & 0.493 & \multicolumn{1}{c|}{0.407} & \textbf{0.700} & \textbf{0.741} & \textbf{0.470} & \multicolumn{1}{c|}{0.508} & 0.572 & 0.590 & 0.396 & \multicolumn{1}{c|}{0.533} & 0.614 \\
\#10 & 0.894 & 0.616 & 0.661 & \multicolumn{1}{c|}{0.660} & 0.899 & 0.658 & 0.683 & \multicolumn{1}{c|}{0.693} & 0.584 & 0.620 & 0.498 & \multicolumn{1}{c|}{0.433} & 0.671 & 0.700 & 0.458 & \multicolumn{1}{c|}{0.524} & 0.591 & 0.621 & 0.425 & \multicolumn{1}{c|}{0.518} & 0.620 \\
\#11 & 0.896 & 0.605 & 0.627 & \multicolumn{1}{c|}{0.658} & 0.908 & 0.637 & 0.651 & \multicolumn{1}{c|}{0.688} & 0.585 & 0.610 & 0.458 & \multicolumn{1}{c|}{0.428} & 0.658 & 0.674 & 0.448 & \multicolumn{1}{c|}{0.497} & 0.577 & 0.605 & 0.410 & \multicolumn{1}{c|}{0.497} & 0.606 \\
Ours & \textbf{0.902} & \textbf{0.624} & \textbf{0.698} & \multicolumn{1}{c|}{0.662} & \textbf{0.909} & \textbf{0.666} & \textbf{0.729} & \multicolumn{1}{c|}{\textbf{0.695}} & \textbf{0.593} & \textbf{0.625} & \textbf{0.506} & \multicolumn{1}{c|}{0.443} & 0.688 & 0.712 & 0.460 & \multicolumn{1}{c|}{0.522} & \textbf{0.594} & 0.621 & 0.421 & \multicolumn{1}{c|}{\textbf{0.539}} & \textbf{0.630} \\ \bottomrule[1pt]%
\end{tabular}
\label{tab: Ablation study}
\end{table*}

\begin{figure}[h]
    \centering
    \captionsetup[subfloat]{labelsep=none,format=plain,labelformat=empty}
    \subfloat[\normalsize (a)]{
		\includegraphics[scale=0.112]{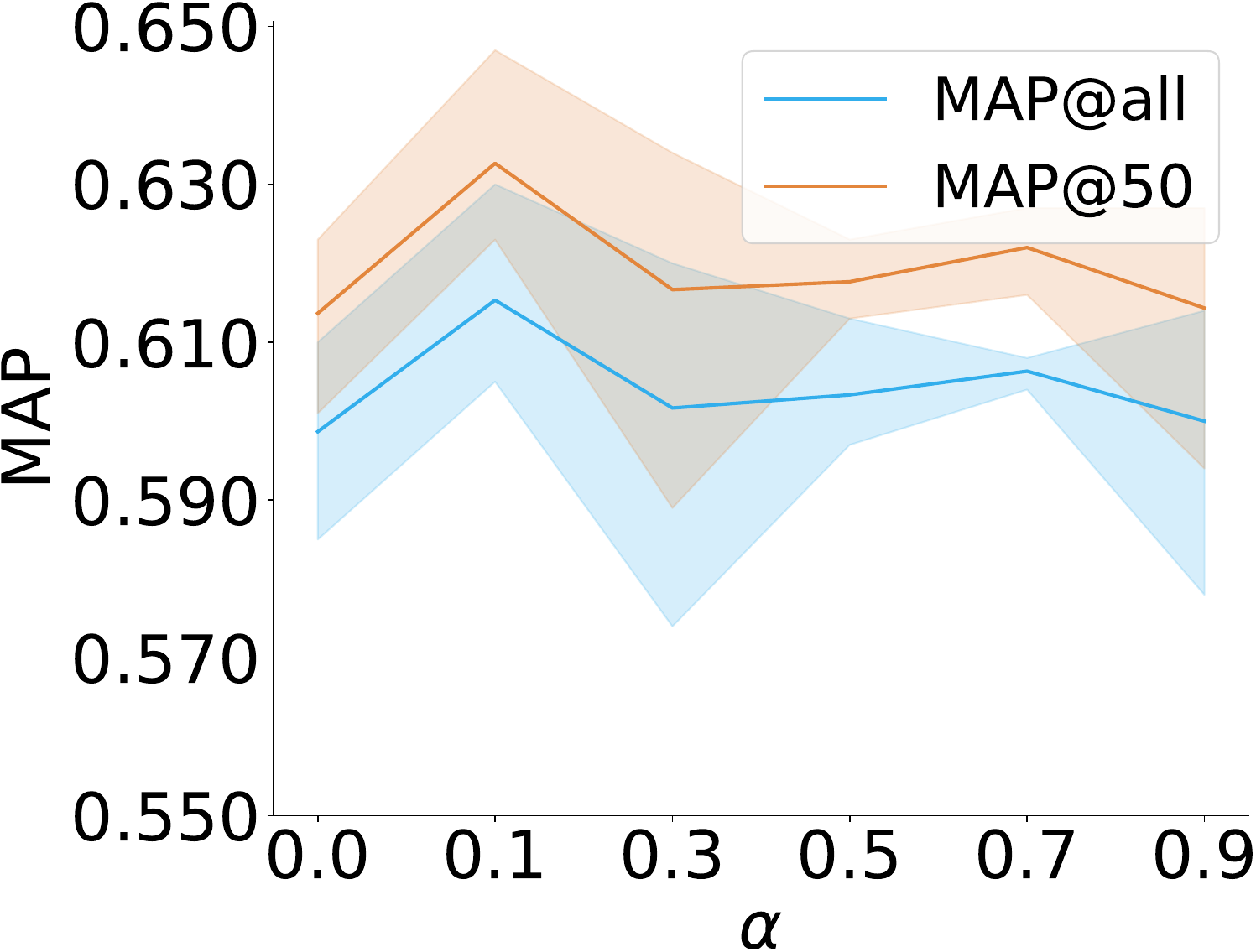}}
    \subfloat[\normalsize (b)]{
		\includegraphics[scale=0.112]{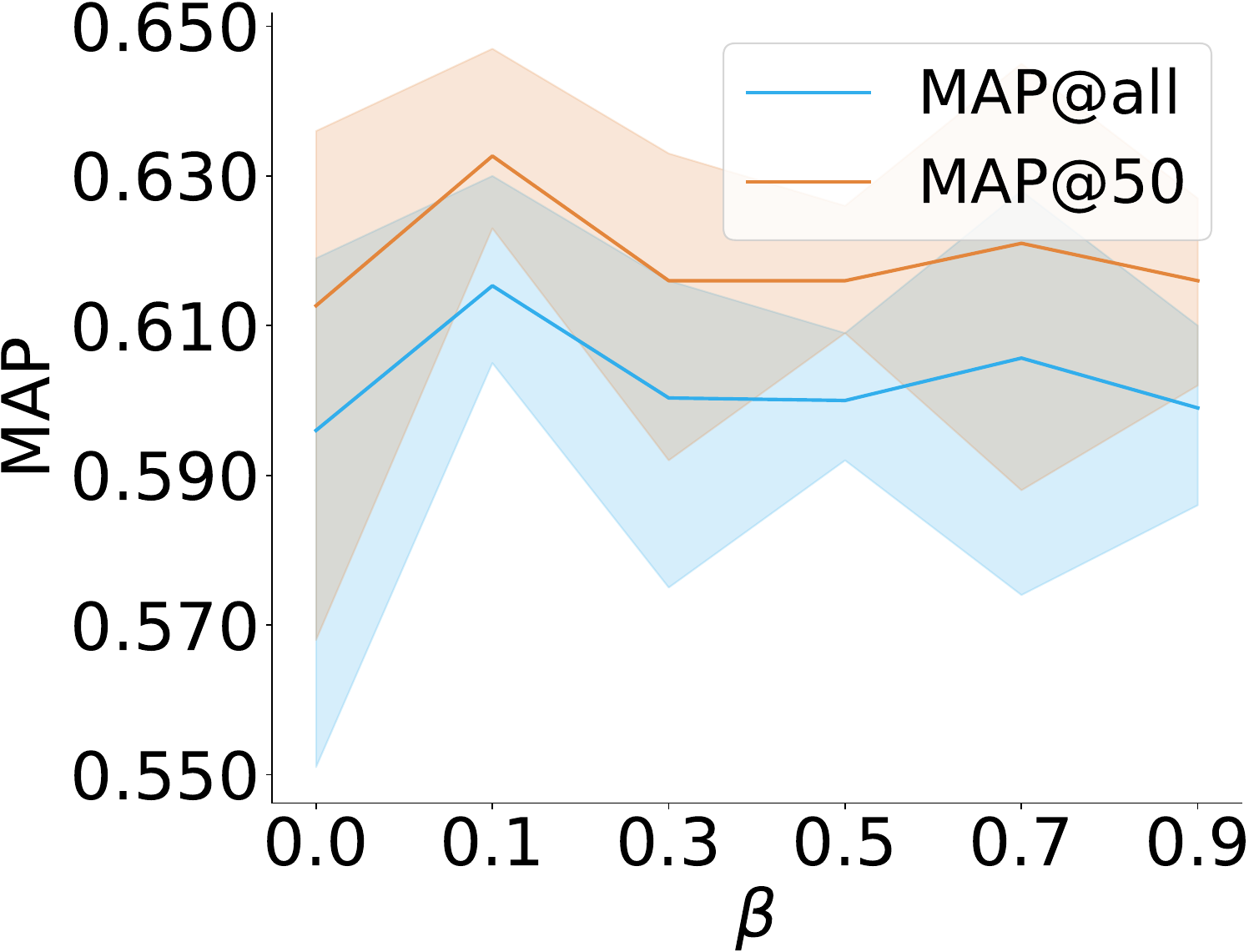}}
    \subfloat[\normalsize (c)]{ 
		\includegraphics[scale=0.112]{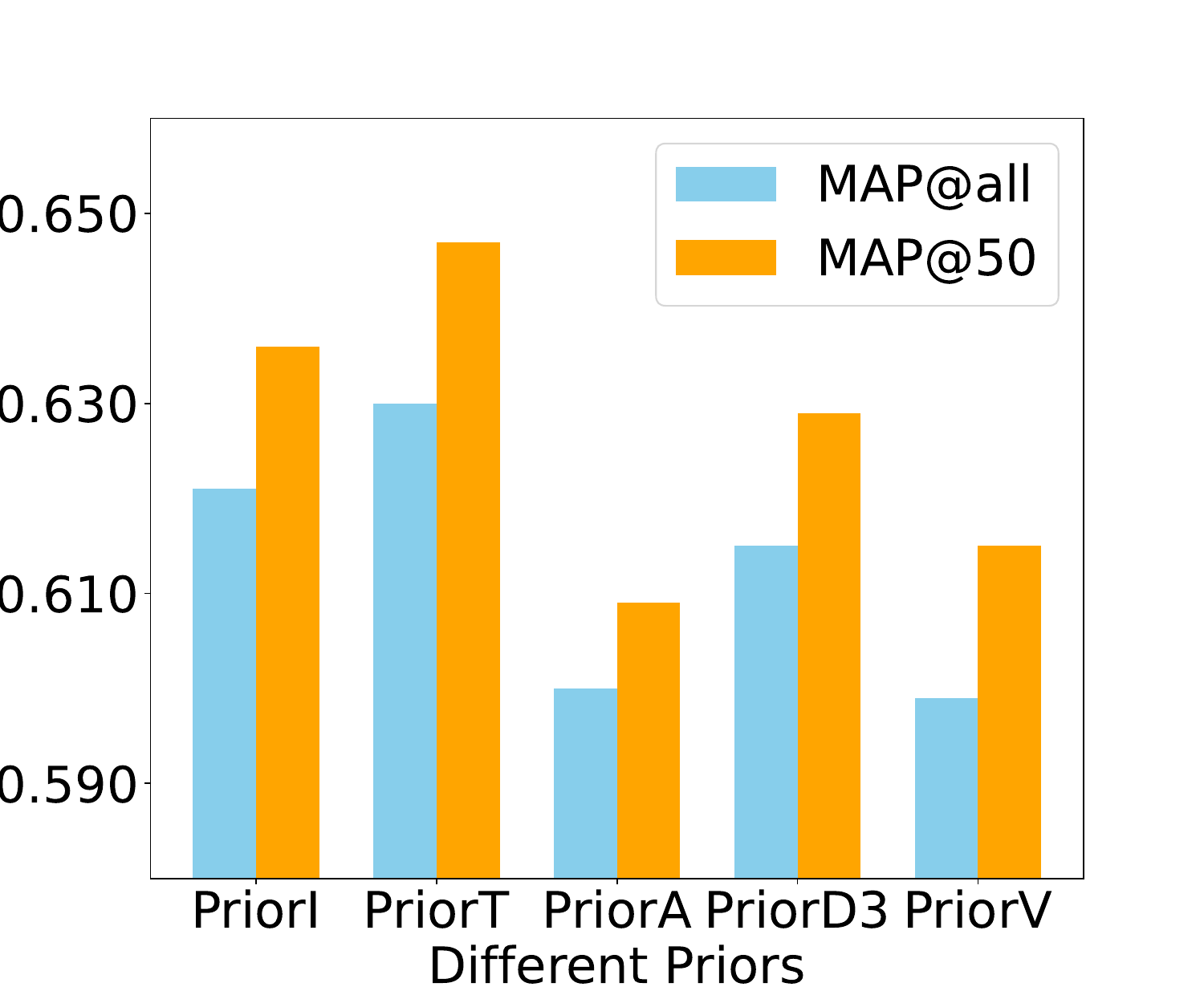}}
  \caption{(a-b) The comparison of MAP@50 and MAP@all scores on the XMedia dataset with different values of hyperparameters $\alpha$ and $\beta$. (c) the comparison of MAP@50 and MAP@all scores on the XMedia dataset with different priors.}
  \label{fig3}
\end{figure} 

\subsection{Ablation Study \label{sec: Ablation-Study}}
In this section, we study the contribution of individual components of DRCL by conducting experiments on 11 variants. Specifically, 1) \#1: Removing SPL and directly using a randomly generated matrix to optimize the model. 2) \#2-4: Removing one of the three losses in RSC (\ie $\mathcal{J}_L$, $\mathcal{J}_{MSE}$, and $\mathcal{J}_{D}$) respectively to optimize the model. 3) \#5-8: Setting the $q$ in $\mathcal{J}_L$ to different fixed values (\ie 0.01, 0.50, 1.00, and 2.00) to optimize the model. 4) \#9: Replacing $\mathbf{W}^{-1}$ with $\mathbf{W}^\top$ to optimize the model. 5) \#10: Removing FA to optimize the model. 6) \#11: Replacing FA with a variant of FA that performs Mixup in feature space rather than embedding space to optimize the model. The detailed settings of the ablation experiment are provided in~\Cref{tab: Ablation study setting} and the MAP results are presented in~\Cref{tab: Ablation study}. It is worth noting that only variant \#1 is designed to study the contribution of SPL, while the other variants are designed to study the contribution of components in RSC. From these results, it can be demonstrated that any removal or alteration of these components adversely impacts the original retrieval performance, affirming that each current component is most beneficial to improving the retrieval performance of the proposed DRCL.

\subsection{Hyperparameter Analysis}\label{sec_hya}
In this section, we study the impact of varying values of the two hyperparameters $\alpha$ and $\beta$ on retrieval performance. We alter the value of one hyperparameter while keeping the other constant at $0.1$. We report the average MAPs of the testing results for models trained with three different random seeds to ensure the persuasiveness of the experiments. The experimental results are depicted in~\Cref{fig3}(a,b). We can observe that the curves in the figures show an unimodal trend, with the optimal performance achieved when both hyperparameters are set to $\alpha=0.1$ and $\beta=0.1$. This configuration is selected as the optimal value for the two hyperparameters in our experiments.

\begin{table}[]
    \centering
    \setlength{\tabcolsep}{1.5pt}
    \caption{The impact of our prior on MAP@all scores on the Wikipedia and XMedia datasets. The abbreviations of `A' and `D' stand for Audio and 3D, respectively. `Avg' means the average MAP score of bidirectional retrieval tasks on two modalities. `Avg*' means the average MAP score of cross-model retrieval tasks on all five modalities.} 
    \begin{tabular}{l|ccc|cccc}
\toprule[1pt]%
\multirow{2}{*}{} & \multicolumn{3}{c|}{Wikipedia} & \multicolumn{4}{c}{XMedia} \\ \cline{2-8} 
 & I→T & \multicolumn{1}{c|}{T →I} & Avg & A→D & \multicolumn{1}{c|}{D→A} & Avg & Avg* \\ \hline
SDML($\mathbf{W_{\text{random}}}$) & 0.545 & \multicolumn{1}{c|}{0.503} & 0.524 & 0.436 & \multicolumn{1}{c|}{0.383} & 0.410 & 0.584 \\
SDML($\mathbf{W_{\text{prior}}}$) & 0.545 & \multicolumn{1}{c|}{0.506} & 0.526 & 0.529 & \multicolumn{1}{c|}{0.430} & 0.480 & 0.608 \\
MARS($\mathbf{W_{\text{random}}}$) & 0.531 & \multicolumn{1}{c|}{0.479} & 0.505 & 0.454 & \multicolumn{1}{c|}{0.430} & 0.442 & 0.590 \\
MARS($\mathbf{W_{\text{prior}}}$) & 0.541 & \multicolumn{1}{c|}{0.504} & 0.523 & 0.494 & \multicolumn{1}{c|}{0.436} & 0.465 & 0.608 \\
DRCL($\mathbf{W_{\text{random}}}$) & 0.544 & \multicolumn{1}{c|}{0.521} & 0.533 & 0.490 & \multicolumn{1}{c|}{0.444} & 0.467 & 0.610 \\
DRCL($\mathbf{W_{\text{prior}}}$) & 0.562 & \multicolumn{1}{c|}{0.537} & 0.550 & 0.506 & \multicolumn{1}{c|}{0.460} & 0.483 & 0.630 \\ \bottomrule[1pt]%
\end{tabular}
    \label{tab: The Impact Of Our Prior}
\end{table}

\subsection{Prior Analysis \label{sec_pa}}
\subsubsection{The impact of different priors}
In our experiments, priors are individually learned on different modalities, followed by the selection of the prior with the best quality score, which is called SPL. In this section, to verify the effectiveness of the prior we selected, we used the prior of each modality to conduct experiments. The experimental results are presented in~\Cref{fig3}(c). We utilize `PriorI', `PriorT', `PriorA', `PriorD3', and `PriorV' to represent the priors learned in images, texts, audio clips, 3D modality, and videos, respectively.  From the results, we can observe that the performance of the text-learned prior surpasses that of the counterparts from other modalities, which affirms the necessity of selecting the most suitable prior.

 \subsubsection{The impact of SPL on other methods} \label{section: The Impact Of Our Prior On Other Methods}
As previously mentioned, some methods (\eg SDML and MARS) employ a random orthogonal matrix for guiding training, neglecting dependencies between inter-class samples. In this section, to further validate the effectiveness of our learned prior, we employ it to guide the training of the aforementioned method. As presented in~\Cref{tab: The Impact Of Our Prior}, on the XMedia dataset, we can observe notable improvements in the average retrieval performance of SDML and MARS, each showing an enhancement of approximately 2\%, with the most substantial improvement observed between audio clips and 3D models. As for the Wikipedia dataset, a discernible performance boost is observed for MARS and DRCL after deploying our prior. This signifies the superior utility of our SPL in guiding model training.

\section{Conclusion}
In this work, we propose a novel approach designed for cross-modal retrieval termed DRCL. DRCL comprises two key modules: the Selective Prior Learning module (SPL) and the Reversible Semantic Consistency learning module (RSC). SPL involves training the shared transformation weight matrix on each modality and selecting the best matrix based on the quality score as the prior. RSC exploits the prior to obtain modality-invariant representations and then uses them to learn the discriminability between inter-class samples while maintaining consistency between label space and common space. The effectiveness and advantage of our method are demonstrated through experimental comparisons with 15 state-of-the-art methods on five widely used datasets.

\textbf{Acknowledgments.} This work is supported in part by the National Natural Science Foundation of China under Grants No. 62372315, and Sichuan Science and Technology Planning Project under Grants No. 2024NSFTD0049, 2024ZDZX0004, 2024YFHZ0144, 2024YFHZ0089, and the Australian Research Council under Grant LP190100594, and the Sichuan Science and Technology Miaozi Program under Grant no. MZGC20240057.
\bibliographystyle{IEEEtran}
\bibliography{main}

\begin{thebibliography}{10}
\providecommand{\url}[1]{#1}
\csname url@samestyle\endcsname
\providecommand{\newblock}{\relax}
\providecommand{\bibinfo}[2]{#2}
\providecommand{\BIBentrySTDinterwordspacing}{\spaceskip=0pt\relax}
\providecommand{\BIBentryALTinterwordstretchfactor}{4}
\providecommand{\BIBentryALTinterwordspacing}{\spaceskip=\fontdimen2\font plus
\BIBentryALTinterwordstretchfactor\fontdimen3\font minus \fontdimen4\font\relax}
\providecommand{\BIBforeignlanguage}[2]{{%
\expandafter\ifx\csname l@#1\endcsname\relax
\typeout{** WARNING: IEEEtran.bst: No hyphenation pattern has been}%
\typeout{** loaded for the language `#1'. Using the pattern for}%
\typeout{** the default language instead.}%
\else
\language=\csname l@#1\endcsname
\fi
#2}}
\providecommand{\BIBdecl}{\relax}
\BIBdecl

\bibitem{NUS-WIDE}
T.-S. Chua, J.~Tang, R.~Hong, H.~Li, Z.~Luo, and Y.~Zheng, ``Nus-wide: a real-world web image database from national university of singapore,'' in \emph{Proceedings of the ACM international conference on image and video retrieval}, 2009, pp. 1--9.

\bibitem{sun2023hierarchical_uni}
Y.~Sun, X.~Wang, D.~Peng, Z.~Ren, and X.~Shen, ``Hierarchical hashing learning for image set classification,'' \emph{IEEE Transactions on Image Processing}, vol.~32, pp. 1732--1744, 2023.

\bibitem{sun2024robust}
Y.~Sun, Y.~Qin, Y.~Li, D.~Peng, X.~Peng, and P.~Hu, ``Robust multi-view clustering with noisy correspondence,'' \emph{IEEE Transactions on Knowledge and Data Engineering}, 2024.

\bibitem{KCCA}
S.~Akaho, ``A kernel method for canonical correlation analysis,'' \emph{arXiv preprint cs/0609071}, 2006.

\bibitem{CCA}
H.~Hotelling, ``Relations between two sets of variates,'' in \emph{Breakthroughs in statistics: methodology and distribution}.\hskip 1em plus 0.5em minus 0.4em\relax Springer, 1992, pp. 162--190.

\bibitem{sun2024dual}
Y.~Sun, J.~Dai, Z.~Ren, Y.~Chen, D.~Peng, and P.~Hu, ``Dual self-paced cross-modal hashing,'' in \emph{Proceedings of the AAAI Conference on Artificial Intelligence}, vol.~38, no.~14, 2024, pp. 15\,184--15\,192.

\bibitem{sun2023hierarchical_bi}
Y.~Sun, Z.~Ren, P.~Hu, D.~Peng, and X.~Wang, ``Hierarchical consensus hashing for cross-modal retrieval,'' \emph{IEEE Transactions on Multimedia}, vol.~26, pp. 824--836, 2023.

\bibitem{NACON}
Y.~Bin, W.~Shi, J.~Zhang, Y.~Ding, Y.~Yang, and H.~T. Shen, ``Non-autoregressive cross-modal coherence modelling,'' in \emph{Proceedings of the 30th ACM International Conference on Multimedia}, 2022, pp. 3253--3261.

\bibitem{HAT}
Y.~Bin, H.~Li, Y.~Xu, X.~Xu, Y.~Yang, and H.~T. Shen, ``Unifying two-stream encoders with transformers for cross-modal retrieval,'' in \emph{Proceedings of the 31st ACM International Conference on Multimedia}, 2023, pp. 3041--3050.

\bibitem{HGAN}
J.~Guo, M.~Wang, Y.~Zhou, B.~Song, Y.~Chi, W.~Fan, and J.~Chang, ``Hgan: Hierarchical graph alignment network for image-text retrieval,'' \emph{IEEE Transactions on Multimedia}, vol.~25, pp. 9189--9202, 2023.

\bibitem{MRL}
P.~Hu, X.~Peng, H.~Zhu, L.~Zhen, and J.~Lin, ``Learning cross-modal retrieval with noisy labels,'' in \emph{Proceedings of the IEEE/CVF conference on computer vision and pattern recognition}, 2021, pp. 5403--5413.

\bibitem{MAN}
P.~Hu, D.~Peng, X.~Wang, and Y.~Xiang, ``Multimodal adversarial network for cross-modal retrieval,'' \emph{Knowledge-Based Systems}, vol. 180, pp. 38--50, 2019.

\bibitem{SDML}
P.~Hu, L.~Zhen, D.~Peng, and P.~Liu, ``Scalable deep multimodal learning for cross-modal retrieval,'' in \emph{Proceedings of the 42nd international ACM SIGIR conference on research and development in information retrieval}, 2019, pp. 635--644.

\bibitem{MARS}
Y.~Wang and Y.~Peng, ``Mars: Learning modality-agnostic representation for scalable cross-media retrieval,'' \emph{IEEE Transactions on Circuits and Systems for Video Technology}, vol.~32, no.~7, pp. 4765--4777, 2021.

\bibitem{ComqueryFormer}
Y.~Xu, Y.~Bin, J.~Wei, Y.~Yang, G.~Wang, and H.~T. Shen, ``Multi-modal transformer with global-local alignment for composed query image retrieval,'' \emph{IEEE Transactions on Multimedia}, vol.~25, pp. 8346--8357, 2023.

\bibitem{DSCMR}
L.~Zhen, P.~Hu, X.~Wang, and D.~Peng, ``Deep supervised cross-modal retrieval,'' in \emph{Proceedings of the IEEE/CVF conference on computer vision and pattern recognition}, 2019, pp. 10\,394--10\,403.

\bibitem{GNN4CMR}
S.~Qian, D.~Xue, Q.~Fang, and C.~Xu, ``Integrating multi-label contrastive learning with dual adversarial graph neural networks for cross-modal retrieval,'' \emph{IEEE Transactions on Pattern Analysis and Machine Intelligence}, vol.~45, no.~4, pp. 4794--4811, 2022.

\bibitem{jSPSH}
Z.~Hu, Y.-M. Cheung, M.~Li, W.~Lan, D.~Zhang, and Q.~Liu, ``Joint semantic preserving sparse hashing for cross-modal retrieval,'' \emph{IEEE Transactions on Circuits and Systems for Video Technology}, vol.~34, no.~4, pp. 2989--3002, 2024.

\bibitem{10684088}
Y.~Qin, C.~Qin, X.~Zhang, and G.~Feng, ``Dual consensus anchor learning for fast multi-view clustering,'' \emph{IEEE Transactions on Image Processing}, vol.~33, pp. 5298--5311, 2024.

\bibitem{9762016}
Y.~Qin, X.~Zhang, L.~Shen, and G.~Feng, ``Maximum block energy guided robust subspace clustering,'' \emph{IEEE Transactions on Pattern Analysis and Machine Intelligence}, vol.~45, no.~2, pp. 2652--2659, 2023.

\bibitem{ASFOH}
Z.~Yang, X.~Deng, L.~Guo, and J.~Long, ``Asymmetric supervised fusion-oriented hashing for cross-modal retrieval,'' \emph{IEEE Transactions on Cybernetics}, vol.~54, no.~2, pp. 851--864, 2023.

\bibitem{MCCA_set}
A.~A. Nielsen, ``Multiset canonical correlations analysis and multispectral, truly multitemporal remote sensing data,'' \emph{IEEE transactions on image processing}, vol.~11, no.~3, pp. 293--305, 2002.

\bibitem{CMCP}
X.~Zhai, Y.~Peng, and J.~Xiao, ``Cross-modality correlation propagation for cross-media retrieval,'' in \emph{2012 IEEE International Conference on Acoustics, Speech and Signal Processing (ICASSP)}.\hskip 1em plus 0.5em minus 0.4em\relax IEEE, 2012, pp. 2337--2340.

\bibitem{GMA}
A.~Sharma, A.~Kumar, H.~Daume, and D.~W. Jacobs, ``Generalized multiview analysis: A discriminative latent space,'' in \emph{2012 IEEE conference on computer vision and pattern recognition}.\hskip 1em plus 0.5em minus 0.4em\relax IEEE, 2012, pp. 2160--2167.

\bibitem{CMSC-DCCA}
Q.~Gao, H.~Lian, Q.~Wang, and G.~Sun, ``Cross-modal subspace clustering via deep canonical correlation analysis,'' in \emph{Proceedings of the AAAI Conference on artificial intelligence}, vol.~34, no.~04, 2020, pp. 3938--3945.

\bibitem{MVMLCCA}
R.~Sanghavi and Y.~Verma, ``Multi-view multi-label canonical correlation analysis for cross-modal matching and retrieval,'' in \emph{Proceedings of the IEEE/CVF conference on computer vision and pattern recognition}, 2022, pp. 4701--4710.

\bibitem{VAE-CCA}
J.~Zhang, Y.~Yu, S.~Tang, J.~Wu, and W.~Li, ``Variational autoencoder with cca for audio--visual cross-modal retrieval,'' \emph{ACM Transactions on Multimedia Computing, Communications and Applications}, vol.~19, no.~3s, pp. 1--21, 2023.

\bibitem{hu2023cross}
P.~Hu, Z.~Huang, D.~Peng, X.~Wang, and X.~Peng, ``Cross-modal retrieval with partially mismatched pairs,'' \emph{IEEE Transactions on Pattern Analysis and Machine Intelligence}, vol.~45, no.~8, pp. 9595--9610, 2023.

\bibitem{ODmAP}
J.~M. Kim, A.~Koepke, C.~Schmid, and Z.~Akata, ``Exposing and mitigating spurious correlations for cross-modal retrieval,'' in \emph{Proceedings of the IEEE/CVF Conference on Computer Vision and Pattern Recognition}, 2023, pp. 2585--2595.

\bibitem{FNE}
H.~Li, Y.~Bin, J.~Liao, Y.~Yang, and H.~T. Shen, ``Your negative may not be true negative: Boosting image-text matching with false negative elimination,'' in \emph{Proceedings of the 31st ACM International Conference on Multimedia}, 2023, pp. 924--934.

\bibitem{CKDH}
J.~Li, W.~K. Wong, L.~Jiang, X.~Fang, S.~Xie, and Y.~Xu, ``Ckdh: Clip-based knowledge distillation hashing for cross-modal retrieval,'' \emph{IEEE Transactions on Circuits and Systems for Video Technology}, vol.~34, no.~7, pp. 6530--6541, 2024.

\bibitem{MLLM}
Y.~Li, W.~Wang, L.~Qu, L.~Nie, W.~Li, and T.-S. Chua, ``Generative cross-modal retrieval: Memorizing images in multimodal language models for retrieval and beyond,'' \emph{arXiv preprint arXiv:2402.10805}, 2024.

\bibitem{radford2021learning}
A.~Radford, J.~W. Kim, C.~Hallacy, A.~Ramesh, G.~Goh, S.~Agarwal, G.~Sastry, A.~Askell, P.~Mishkin, J.~Clark \emph{et~al.}, ``Learning transferable visual models from natural language supervision,'' in \emph{International conference on machine learning}.\hskip 1em plus 0.5em minus 0.4em\relax PMLR, 2021, pp. 8748--8763.

\bibitem{CL2CM}
Y.~Wang, F.~Wang, J.~Dong, and H.~Luo, ``Cl2cm: Improving cross-lingual cross-modal retrieval via cross-lingual knowledge transfer,'' in \emph{Proceedings of the AAAI Conference on Artificial Intelligence}, vol.~38, no.~6, 2024, pp. 5651--5659.

\bibitem{AlRet}
Y.~Xu, Y.~Bin, J.~Wei, Y.~Yang, G.~Wang, and H.~T. Shen, ``Align and retrieve: Composition and decomposition learning in image retrieval with text feedback,'' \emph{IEEE Transactions on Multimedia}, 2024.

\bibitem{hu2022unsupervised}
P.~Hu, H.~Zhu, J.~Lin, D.~Peng, Y.-P. Zhao, and X.~Peng, ``Unsupervised contrastive cross-modal hashing,'' \emph{IEEE Transactions on Pattern Analysis and Machine Intelligence}, vol.~45, no.~3, pp. 3877--3889, 2022.

\bibitem{hu2023deep}
P.~Hu, L.~Zhen, X.~Peng, H.~Zhu, J.~Lin, X.~Wang, and D.~Peng, ``Deep supervised multi-view learning with graph priors,'' \emph{IEEE Transactions on Image Processing}, vol.~33, pp. 123--133, 2023.

\bibitem{COXI}
Y.~Wei, L.~Zheng, G.~Qiu, and G.~Cai, ``Cross-modal retrieval based on shared proxies,'' \emph{International Journal of Multimedia Information Retrieval}, vol.~13, no.~1, pp. 1--16, 2024.

\bibitem{ELRCMR}
T.~Xu, X.~Liu, Z.~Huang, D.~Guo, R.~Hong, and M.~Wang, ``Early-learning regularized contrastive learning for cross-modal retrieval with noisy labels,'' in \emph{Proceedings of the 30th ACM International Conference on Multimedia}, 2022, pp. 629--637.

\bibitem{zhang2017mixup}
H.~Zhang, M.~Cisse, Y.~N. Dauphin, and D.~Lopez-Paz, ``mixup: Beyond empirical risk minimization,'' \emph{arXiv preprint arXiv:1710.09412}, 2017.

\bibitem{2024mixup-cross-modal-retrieval}
F.~Ding, X.~Liu, X.~Wang, and F.~Zhong, ``Dual-mix for cross-modal retrieval with noisy labels,'' in \emph{ICASSP 2024-2024 IEEE International Conference on Acoustics, Speech and Signal Processing (ICASSP)}.\hskip 1em plus 0.5em minus 0.4em\relax IEEE, 2024, pp. 6505--6509.

\bibitem{Xmedia}
Y.~Peng, X.~Zhai, Y.~Zhao, and X.~Huang, ``Semi-supervised cross-media feature learning with unified patch graph regularization,'' \emph{IEEE transactions on circuits and systems for video technology}, vol.~26, no.~3, pp. 583--596, 2015.

\bibitem{XmediaNet}
Y.~Peng, X.~Huang, and Y.~Zhao, ``An overview of cross-media retrieval: Concepts, methodologies, benchmarks, and challenges,'' \emph{IEEE Transactions on circuits and systems for video technology}, vol.~28, no.~9, pp. 2372--2385, 2017.

\bibitem{Wiki}
N.~Rasiwasia, J.~Costa~Pereira, E.~Coviello, G.~Doyle, G.~R. Lanckriet, R.~Levy, and N.~Vasconcelos, ``A new approach to cross-modal multimedia retrieval,'' in \emph{Proceedings of the 18th ACM international conference on Multimedia}, 2010, pp. 251--260.

\bibitem{INRIA-Websearch}
J.~Krapac, M.~Allan, J.~Verbeek, and F.~Juried, ``Improving web image search results using query-relative classifiers,'' in \emph{2010 IEEE Computer Society Conference on Computer Vision and Pattern Recognition}.\hskip 1em plus 0.5em minus 0.4em\relax IEEE, 2010, pp. 1094--1101.

\bibitem{ACMR}
B.~Wang, Y.~Yang, X.~Xu, A.~Hanjalic, and H.~T. Shen, ``Adversarial cross-modal retrieval,'' in \emph{Proceedings of the 25th ACM international conference on Multimedia}, 2017, pp. 154--162.

\bibitem{mmsae}
Y.~Wu, S.~Wang, and Q.~Huang, ``Multi-modal semantic autoencoder for cross-modal retrieval,'' \emph{Neurocomputing}, vol. 331, pp. 165--175, 2019.

\bibitem{DRSL}
X.~Wang, P.~Hu, L.~Zhen, and D.~Peng, ``Drsl: Deep relational similarity learning for cross-modal retrieval,'' \emph{Information Sciences}, vol. 546, pp. 298--311, 2021.

\bibitem{ALGCN}
S.~Qian, D.~Xue, Q.~Fang, and C.~Xu, ``Adaptive label-aware graph convolutional networks for cross-modal retrieval,'' \emph{IEEE Transactions on Multimedia}, vol.~24, pp. 3520--3532, 2021.

\bibitem{RONO}
Y.~Feng, H.~Zhu, D.~Peng, X.~Peng, and P.~Hu, ``Rono: Robust discriminative learning with noisy labels for 2d-3d cross-modal retrieval,'' in \emph{Proceedings of the IEEE/CVF Conference on Computer Vision and Pattern Recognition}, 2023, pp. 11\,610--11\,619.

\end{thebibliography}

\end{document}